\documentclass[runningheads]{llncs}
\usepackage{graphicx}

\usepackage{tikz}
\usepackage{comment}
\usepackage{amsmath,amssymb} 
\usepackage{color}
\usepackage{booktabs}
\usepackage{bm}
\usepackage{caption}
\usepackage{amsfonts}
\usepackage{times}
\usepackage{epsfig}
\usepackage{multirow}
\usepackage{relsize}
\usepackage{verbatim}
\usepackage{subcaption}
\usepackage{wrapfig}
\usepackage{array}
\newcolumntype{P}[1]{>{\centering\arraybackslash}p{#1}}

\newcommand{\etal}{et al.}

\begin{document}
\title{ConTra: (Con)text (Tra)nsformer\\for Cross-Modal Video Retrieval}
\titlerunning{ConTra for Cross-Modal Video Retrieval}
%
\author{Adriano Fragomeni \qquad \quad Michael Wray \qquad \quad Dima Damen}
\authorrunning{A. Fragomeni \etal}
%
\institute{Department of Computer Science, University of Bristol, UK}
\maketitle              
%

\begin{abstract}
In this paper, we re-examine the task of cross-modal clip-sentence retrieval, where the clip is part of a longer untrimmed video. 
When the clip is short or visually ambiguous, knowledge of its local temporal context (i.e. surrounding video segments) can be used to improve the retrieval performance.
We propose \textbf{Con}text \textbf{Tra}nsformer (ConTra); an encoder architecture that models the interaction between a video clip  and its local temporal context in order to enhance its embedded representations.
Importantly, we supervise the context transformer using contrastive losses in the  cross-modal embedding space.

We explore context transformers for video and text modalities.
Results consistently demonstrate improved performance on three datasets: YouCook2, EPIC-KITCHENS and a clip-sentence version of ActivityNet Captions. Exhaustive ablation studies and context analysis show the efficacy of the proposed method.
\end{abstract}

\section{Introduction}
\label{sec:intro}

Millions of hours of video are being uploaded to online platforms every day. 
Leveraging this wealth of visual knowledge relies on methods that can understand the video, whilst also allowing for videos to be searchable, e.g. via language.
Methods can query the entire video~\cite{DBLP:conf/eccv/ZhangHS18,DBLP:conf/eccv/ShaoXZHQL18,DBLP:conf/nips/GingZPB20} or the individual segments, or clips, that make up a video~\cite{dong2016word2visualvec,DBLP:journals/corr/abs-1804-02516}.
In this work, we focus on the latter problem of clip-sentence retrieval, specifically from long untrimmed videos.
This is particularly beneficial to retrieve all instances of the same step (e.g. folding dough or jacking up a car) from videos of various procedures.

\begin{figure}[t!]
\begin{center}
\includegraphics[width=\linewidth]{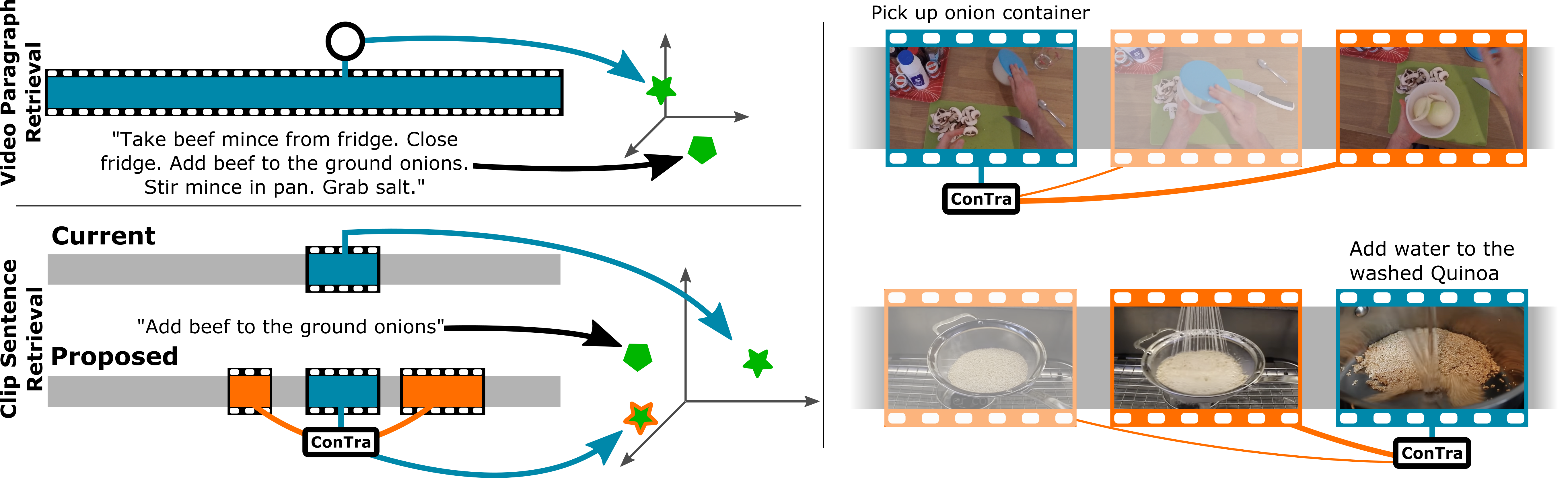}
\end{center}
\vspace*{-14pt}
\caption{
\textbf{Left}: We compare video-paragraph retrieval (top) to current and proposed clip-sentence retrieval (bottom) in long videos. In ConTra, we propose to attend to local context of neighbouring clips.
\textbf{Right}: Examples where ConTra can enrich the clip representation from next/previous clips, observing the onion (top) or that the quinoa has already been washed (bottom). Line thickness/brightness represents attention weights.}
\vspace*{-15pt}
\label{fig:intro_fig}
\end{figure}

In Fig.~\ref{fig:intro_fig}, we compare current clip-sentence retrieval approaches  (e.g.~\cite{DBLP:conf/iclr/Patrick0AMHHV21,DBLP:conf/cvpr/WangZ021,DBLP:journals/corr/abs-2104-08271,DBLP:journals/corr/abs-2104-12671,DBLP:conf/cvpr/MiechALSZ21,DBLP:journals/corr/abs-2104-00650}) to our proposed context transformer. 
We leverage local temporal context clues, readily available in long videos, to improve retrieval performance. 
Local sequences of actions often use similar objects or include actions towards the same goal, which can enrich the embedded clip representation. 

We emphasise the importance of learnt
\textbf{local temporal \mbox{clip context}}, that is the \emph{few} clips surrounding (i.e. before and after) the clip to be embedded.
Our model, ConTra, learns to attend to relevant neighbouring clips by using a transformer encoder, differing from previous works which learn context over frames~\cite{DBLP:conf/eccv/Gabeur0AS20} or globally across the entire video~\cite{DBLP:conf/nips/GingZPB20} (see Video-Paragraph Retrieval, Fig.~\ref{fig:intro_fig} top).
We supervise ConTra with cross-modal contrastive losses and a proposed neighbouring loss that ensures the embedding is distinct across overlapping contexts.

Our contributions are summarised as follows: (i)~we explore the task of cross-modal clip-sentence
retrieval when using local context in clip, text or in both modalities simultaneously (ii)~we propose ConTra, a transformer based architecture that learns to attend to local
temporal context, supervised by a multi-term loss that is able to distinguish consecutive
clips by the introduction of a neighbouring contrastive loss (iii)~we demonstrate the
added value of local context by conducting detailed experiments on three datasets.  

\section{Related works}
\label{sec:related}

In this section, we split video-text retrieval works into those which primarily focus on either Clip-Sentence or Video-Paragraph retrieval before presenting works that use temporal context for other video understanding tasks.

\noindent \textbf{Clip-Sentence Retrieval:} \quad
Most works primarily rely on two-stream (i.e. dual) approaches~\cite{DBLP:journals/corr/abs-2104-00650,DBLP:conf/cvpr/KleinLSW15,DBLP:conf/cvpr/MiechASLSZ20,DBLP:conf/iccv/MiechZATLS19,DBLP:conf/cvpr/WangZ021,DBLP:conf/cvpr/WeiX0JWS20,DBLP:conf/iclr/Patrick0AMHHV21,DBLP:journals/corr/abs-2105-13033,DBLP:journals/corr/abs-2104-00285,xu-etal-2021-videoclip}, using multiple text embeddings~\cite{DBLP:conf/cvpr/ChenZJW20,DBLP:conf/iccv/WrayCLD19,DBLP:journals/corr/abs-2104-08271,Dong_2021,DBLP:conf/mir/MithunLMR18}, video experts~\cite{DBLP:conf/bmvc/LiuANZ19,DBLP:conf/cvpr/0105CA21,DBLP:journals/corr/abs-1804-02516}, or audio~\cite{DBLP:conf/nips/AlayracRSARFSDZ20,DBLP:journals/corr/abs-2104-12671,DBLP:conf/cvpr/WangZ021,DBLP:journals/corr/abs-2104-11178}. Recently, single stream cross-modal encoders have also been used~\cite{DBLP:conf/cvpr/MiechALSZ21,DBLP:conf/cvpr/ZhuY20a,DBLP:conf/naacl/TangLB21,DBLP:journals/corr/abs-2002-06353,DBLP:conf/acl/XuGHAAFMZ21},  
improving inter-modality modelling at the cost of increased computational complexity. In ConTra, we use a dual stream model with separate branches for the visual and the textual components.

Temporal modelling of frames {\it{within a clip}} is a common avenue for retrieval approaches~\cite{DBLP:conf/eccv/Gabeur0AS20,DBLP:conf/cvpr/WangZ021,DBLP:conf/cvpr/ZhuY20a}.
Gabeur~\etal~\cite{DBLP:conf/eccv/Gabeur0AS20} use multiple video experts with a multi-modal transformer to better capture the temporal relationships between modalities. Wang~\etal~\cite{DBLP:conf/cvpr/WangZ021} learn an alignment between words and frame features alongside the clip-sentence alignment.
ActBert~\cite{DBLP:conf/cvpr/ZhuY20a} also models alignment between clip and word-level features using self-supervised learning. 
Bain~\etal~\cite{DBLP:journals/corr/abs-2104-00650} adapt a ViT~\cite{DBLP:conf/iclr/DosovitskiyB0WZ21} model, trained with a curriculum learning schedule,
to gradually attend to more frames within each clip.
MIL-NCE~\cite{DBLP:conf/cvpr/MiechASLSZ20}  alleviates noise within the automated captions, matching clips to neighbouring sentences. However, the learned representation does not go beyond the clip extent.
VideoCLIP~\cite{xu-etal-2021-videoclip} creates positive clips by sampling both the centre-point~(within narration timestamp) and the clip's duration to better align clips and sentences, foregoing the reliance on explicit start/end times.
In our work, we go beyond temporal modelling of the clip itself to using local context outside the clip.

Other works improve modelling of the textual representation~\cite{Dong_2021,DBLP:conf/iccv/WrayCLD19,DBLP:conf/iclr/Patrick0AMHHV21}.
Patrick~\etal~\cite{DBLP:conf/iclr/Patrick0AMHHV21} introduce a generative task of cross-instance captioning to alleviate false negatives. They create a support set of relevant captions and learn to reconstruct a sample’s text representation as a weighted combination of a support-set of video representations from the batch.
However, whilst they use information from other sentences, there is no notion of those which are temporally related.
Instead, we propose to explore relationships between neighbouring sentences using local context within the same video. 

\noindent \textbf{Video-paragraph Retrieval:} \quad
Another retrieval task is video-paragraph retrieval~\cite{DBLP:conf/eccv/ZhangHS18,DBLP:conf/eccv/ShaoXZHQL18,DBLP:conf/cvpr/WeiX0JWS20,DBLP:conf/bmvc/LiuANZ19,DBLP:journals/corr/abs-2104-08271,DBLP:conf/cvpr/WangZ021,DBLP:conf/iccv/SunMV0S19,DBLP:journals/corr/abs-2102-06183,DBLP:conf/nips/GingZPB20}, where videos and paragraphs describing the full videos are embedded in their entirety. There are two main approaches: using  hierarchical representations between the paragraph/video and constituent sentences/clips~\cite{DBLP:conf/eccv/ZhangHS18,DBLP:conf/bmvc/LiuANZ19,DBLP:conf/nips/GingZPB20} or jointly modelling the entire video/paragraph with a cross-modal transformer~\cite{DBLP:conf/iccv/SunMV0S19,DBLP:journals/corr/abs-2102-06183}.

COOT~\cite{DBLP:conf/nips/GingZPB20} models the interactions between levels of granularity for each modality by using a hierarchical transformer.
The video and paragraph embeddings are obtained via a combination of their clips and sentences.
ClipBERT~\cite{DBLP:journals/corr/abs-2102-06183} inputs both text and video to a single transformer encoder after employing sparse sampling, where only a single or a few sampled clips are used at each training step. 
In this work, we focus on clip-sentence retrieval, but take inspiration from video-paragraph works in how they relate clips within a video. Importantly, we focus on local context, which 
is applicable to long videos with hundreds of clips. 

\noindent \textbf{Temporal Context for Video Understanding:} \quad
We also build on works that successfully used \emph{local temporal context} for other video understanding tasks such as: action recognition~\cite{kazakos2021MTCN,DBLP:journals/corr/abs-2104-11520,DBLP:conf/cvpr/WuF0HKG19,DBLP:conf/cvpr/Zhang0Z21}; action anticipation~\cite{DBLP:conf/eccv/SenerSY20,furnari2019would};  object detection and tracking~\cite{DBLP:conf/cvpr/BeeryWRVH20,DBLP:conf/cvpr/BertasiusT20}; moment localisation~\cite{DBLP:journals/tip/ZhangHSYN21}; and Content-Based Retrieval~\cite{DBLP:conf/wacv/ShaoWZX21}. 
Bertasius and Torresani~\cite{DBLP:conf/cvpr/BertasiusT20} use local context for mask propagation to better segment and track occluded objects.
Kazakos~\etal~\cite{kazakos2021MTCN} use the context of neighbouring clips for action recognition using a tranformer encoder along with a language model to ensure the predicted sequence of actions is realistic.
Feichtenhofer~\etal~\cite{DBLP:conf/cvpr/WuF0HKG19} allow for modelling features beyond the short clips.
They use a non-local attention block and find that using context from up to 60 seconds can help recognise actions.
Shao~\etal~\cite{DBLP:conf/wacv/ShaoWZX21} use a self-attention mechanism to model long-term dependencies for content-based video retrieval. They use a supervised contrastive learning method that performs automatic hard negative mining and utilises a memory bank to increase
the capacity of negative samples.

To the best of our knowledge, ours is the first work to explore using neighbouring clips as context for cross-modal clip-sentence retrieval in long untrimmed videos.

\section{Context Transformer (ConTra)}

\begin{figure*}
  \centering
  \includegraphics[width=1\linewidth]{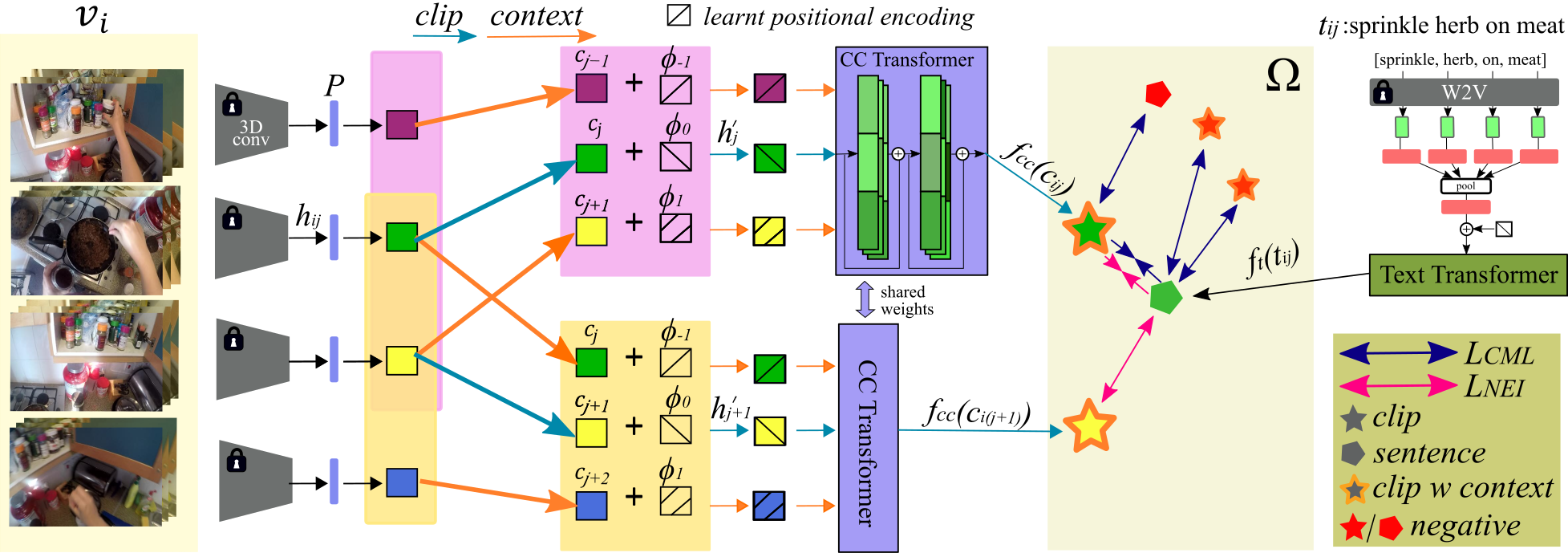}
  \vspace*{-14pt}
  \caption{\textbf{Overview of ConTra:} Given a video $v_i$ split into clips $c_{ij}$, we encode clips into features $h_{ij}$, projected by $P$ and tagged with a learnt position encoding $\phi$ \emph{relative} to the centre clip. A Clip Context (CC) encoder learns an enriched representation of the centre clip (\textcolor{cyan}{cyan arrow}) attending to context clips (\textcolor{orange}{orange arrows}). The embedding space $\Omega$ is learnt with cross-modal (CML) and neighbouring (NEI) losses. NEI pushes overlapping contexts further apart---shown for $f_{cc}(c_{ij})$ and $f_{cc}(c_{i(j+1)})$.}
  \vspace*{-15pt}
  \label{fig:fig2}
\end{figure*}

We first explicitly define the task of clip-sentence retrieval in Sec.~\ref{subsec:method:definition} before extending the definition to incorporate \emph{local} clip context, in untrimmed videos. 
We then present our clip context embedding in Sec.~\ref{subsec:method:contra} where we provide details of our architecture followed by the training losses in Sec.~\ref{subsec:method:losses}.
We then extend ConTra to context in both modalities in Sec.~\ref{sec:method:modality}.
An overview of our approach can be seen in Fig.~\ref{fig:fig2}.
\vspace*{-10pt}
\subsection{Definitions} 
\label{subsec:method:definition}
\vspace*{-6pt}
We begin with a set of untrimmed videos, $v_i \in V$.
These are broken down further into ordered clips, 
$c_{ij} \in v_i$, each with a corresponding sentence/caption, $t_{ij}$, describing the action within the clip. 
 Querying by the sentence $t_{ij}$ aims to retrieve the corresponding clip $c_{ij}$ and vice versa for cross-modal retrieval.

Learning a dual stream retrieval model focuses on learning two projection functions, $f_c : c \longrightarrow \Omega \subseteq \mathbb{R}^d$ and $f_t : t \longrightarrow \Omega \subseteq \mathbb{R}^d$, which project the video/text modalities respectively into a common $d$-dimensional embedding space, $\Omega$, where $c_{ij}$ and $t_{ij}$ are close. The weights of these embedding functions can be collectively trained using contrastive-based losses including a triplet loss~\cite{DBLP:conf/cvpr/WangLL16,DBLP:journals/pami/WangLHL19,DBLP:conf/cvpr/WangSLRWPCW14,DBLP:conf/cvpr/SchroffKP15} or noise-contrastive estimation~\cite{DBLP:journals/jmlr/GutmannH12,DBLP:journals/corr/JozefowiczVSSW16,DBLP:conf/cvpr/MiechASLSZ20}.

Instead of using the clip solely, we wish to utilise its \emph{local} temporal context to enrich the embedded representation.
We define the temporal Clip Context~(CC) using $m$, around the clip $c_{ij}$, as follows:
\begin{equation}
    CC_m(c_{ij}) = (c_{i(j-m)}, \cdots,c_{ij}, \cdots, c_{i(j+m)})
    \label{eqn:clipcontext}
\end{equation}
There are $2m$ clips that are temporally adjacent to $c_{ij}$ in the same video $v_i$. Note that the length of the adjacent clips governed by $m$ differs per dataset and video. 
Importantly, we still aim to retrieve the corresponding sentence $t_{ij}$ for the centre clip $c_{ij}$, but utilise the untrimmed nature of the video around the clip to enrich $c_{ij}$'s representation.

In Table~\ref{tab:settings}, we differentiate between existing tasks in 
Sec.~\ref{sec:related} 
and our 
proposed 
\begin{wraptable}[10]{r}{.6\linewidth}
    \vspace*{-20pt}
    \begin{tabular}{lccc}
                     Task  &  Clip?  & Video & Text\\ \hline
      Video-Paragraph & $\times$     & all clips & all sentences  \\\hline
      Clip-Sentence\\
        \hspace{12pt}No Context      & $\checkmark$ &   clip    &   sentence     \\ 
        \hspace{12pt}Clip Context    & $\checkmark$ & clip+context  &   sentence     \\
        \hspace{12pt}Text Context    & $\checkmark$ &   clip    & sent.+context   \\
        \hspace{12pt}Context in Both     & $\checkmark$ & clip+context  & sent.+context   \\ \hline
    \end{tabular}
    \vspace*{-6pt}
    \caption{Comparison of tasks with/without using context in video and text. }
    \label{tab:settings}
\end{wraptable}
settings.
Note that in Video-Paragraph retrieval, models cannot be used to retrieve individual clips. 
Different from the standard Clip-Sentence setting, we utilise neighbouring clips to enrich the clip representation, the sentence representation or both. 
Next, we describe our Clip Context Transformer.

\subsection{Clip Context Transformer}
\label{subsec:method:contra}
We learn the embedding function $f_{cc}:  CC_m(c) \longrightarrow \Omega$, using the local clip context in Eq.~\ref{eqn:clipcontext}.
We consider each clip as a linear projection of its features $P(h_j)$.
We drop the video index $i$ here for simplicity.
We learn $2m+1$ distinct positional embeddings, $(\phi_{-m}, \cdots, \phi_0, \cdots, \phi_m)$, that are added such that $h_{j+\alpha}' = P(h_{j+\alpha}) + \phi_{0+\alpha}$,  where $-m \le \alpha \le m$ and $\phi_0$ is the positional embedding of the centre clip.
Note that the positional embeddings emphasise the order of the clip within the context window rather than the full video, thus reflecting the \emph{relative} position of the context to the centre clip~$c_{ij}$, and are identical across contexts.
We showcase this on two neighbouring clips in Fig.~\ref{fig:fig2}.
We form the input to the encoder transformer as:
\begin{equation}
    H' = [h_{j-m}', \cdots, h_j', \cdots, h_{j+m}']
\end{equation}

From $H'$, we aim to learn the embedding of the centre clip. We use a multi-headed attention block~\cite{DBLP:journals/corr/VaswaniSPUJGKP17} with the standard self-attention heads and residual connections.
The output of the $r^{th}$ attention head is thus computed as, 
\begin{equation}
    A_{r} = \sigma \left(\frac{(\bm{\theta_r^Q}\bm{H'})(\bm{\theta_r^K}\bm{H'})^\top}{\sqrt{d}}\right) (\bm{\theta_r^W}\bm{H'})
\end{equation}
where $\bm{\theta^Q_r}$, $\bm{\theta^K_r}$, $\bm{\theta^W_r} \in \mathbb{R}^{(2m+1) \times d/R}$ are learnable projection matrices.
The output of the multi-head attention is then calculated as the concatenation of all $R$ heads:
\begin{equation}
    A = [A_{1}, \dots, A_{R}] + H'
\end{equation}
For the clip embedding, we focus on the output from $A$ corresponding to the centre clip $j$, such that:
\begin{equation}
    f_{cc}(c_{ij}) = g(A_j) + A_j
\end{equation}
where $g$ is one or more linear layers with ReLU activations, along with another residual connection.
Note that the size of $f_{cc}$ is $d$, independent of the context length $m$. 
$f_{cc}$ can be extended with further multi-head attention layers.
We discuss how we train the ConTra model next.
\vspace*{-10pt}
\subsection{Training ConTra}
\vspace*{-6pt}
\label{subsec:method:losses}
\paragraph{Cross-Modal Loss.}\quad For both training and inference, we calculate the cosine similarity $s(c_{ij}, t_{kl})$ between the embeddings of the context-enriched clip $f_{cc}(c_{ij})$ and a sentence~$f_t(t_{kl})$.
Cross-modal losses are regularly used in retrieval works such as the triplet loss~\cite{DBLP:conf/nips/GingZPB20,DBLP:conf/iccv/MiechZATLS19,DBLP:conf/cvpr/ChenZJW20,DBLP:conf/bmvc/LiuANZ19} and the Noise-Contrastive Estimation (NCE) loss~\cite{DBLP:journals/corr/abs-2104-12671,DBLP:journals/corr/abs-2104-11178,DBLP:conf/acl/XuGHAAFMZ21,Liu2021HiTHT}. We use NCE as our cross-modal loss~($L_{CML}$)~\cite{DBLP:journals/jmlr/GutmannH12,DBLP:journals/corr/JozefowiczVSSW16}:

\vspace*{-9pt}
\begin{equation}
    L_{CML}= \frac{1}{|B|}\sum\limits_{(c_{ij},t_{ij}) \in B} -\log{\left(\frac{e^{ s(c_{ij}, t_{ij}) / \tau}}{e^{ s(c_{ij}, t_{ij})/ \tau} + \sum\limits_{(c',t') \sim \mathcal{N'}} e^{ s(c', t') / \tau}}\right)}
    \label{eqn:NCE}
\end{equation}

\noindent where 
$B$ is a set of corresponding clip-captions pairs, i.e. $(c_{ij}, t_{ij})$ and $\tau$ is the temperature parameter.
We construct the negative set $\mathcal{N'}$ in each case from the batch by combining $(c_{ij}, t_{lk})_{ij \ne lk}$ as well as $(c_{lk}, t_{ij})_{ij \ne lk}$, considering negatives for both clip and sentence across elements in the batch.

\paragraph{Uniformity Loss.}\quad
The uniformity loss is less regularly used, but was proposed in~\cite{DBLP:conf/icml/0001I20} and used in~\cite{DBLP:conf/cvpr/ChunORKL21,DBLP:journals/corr/abs-2102-05644} works, for image retrieval.
It ensures that the embedded representations preserve maximal information, i.e. feature vectors are distributed uniformly on the unit hypersphere. We use the uniformity loss ($L_{UNI}$) such as:
\begin{equation}
\label{eqn:uniformity}
    L_{UNI}=\log{\left(\frac{1}{|B|}\sum_{u,u' \in U \times U} e^{-2\| u-u'\|_2^2} \right)}
\end{equation}
where $U=\{c_1,t_1,...,c_B,t_B\}$, are all the clips and sentences in the batch. This loss term is
applied to all the clips and sentences in a batch.

\paragraph{Neighbouring Loss.}\quad
We additionally propose a neighbour-contrasting loss~($L_{NEI}$) to ensure that the embeddings of context items are well discriminated. 
Indeed, one of the challenges of introducing local temporal context is the overlap between contexts of neighbouring clips. 
Consider two neighbouring clips in the same video, say $c_{ij}$ and $c_{i(j+1)}$  (see Fig.~\ref{fig:fig2}), the context windows $CC(c_{ij})$ and $CC(c_{i(j+1)})$ share $2m$ clips.
While the positional encoding of the clips differ, distinguishing between the embedded neighbouring clips can be challenging.
This can be considered as a special case of hard negative mining, as in~\cite{DBLP:conf/bmvc/FaghriFKF18,DBLP:conf/emnlp/HendricksWSSDR18}, however
our usage of it, where only neighbouring clips are considered as negatives is novel. 

Accordingly, we define the $L_{NEI}$ using the NCE loss:
\begin{equation}
\label{eqn:Neighbouring}
    L_{NEI}=\mathlarger{\frac{1}{|B|}} \mathlarger{\mathlarger{\sum}}_{(c_{ij},t_{ij})\in B} -\log{\left(\frac{e^{ s(c_{ij}, t_{ij}) / \tau}}{e^{ s(c_{ij}, t_{ij})/ \tau} + e^{ s(c_{i(j+\alpha)}, t_{ij}) / \tau}}\right)}
\end{equation}
where $\alpha$ is randomly sampled from $[-m,  m]$ subject to ${t_{ij} \ne t_{i(j+\alpha)}}$. We thus randomly sample a neighbouring clip, avoiding neighbours where the sentences are matching (e.g. the sentence, ``mix ingredients'' might be repeated in consecutive clips). 

In practice, the neighbouring loss is calculated by having another batch of sampled neighbouring clips of size $B$. We use a single negative neighbour per clip to keep the batch size to $B$ regardless of the length $m$, though we do ablate differing numbers of sampled negatives in Sec.~\ref{sec:exp-clip-context}.



We optimize our ConTra model by minimizing the overall loss function $L$:
\begin{equation}
\label{eqn:total_loss}
    L=L_{CML} + L_{NEI} + L_{UNI}
\end{equation}
We keep the weights between the three losses the same in all experiments and datasets showcasing that we outperform other approaches without hyperparameter tuning.
In supplementary, we report results when tuning the weights, to ablate these.

Once the model is trained, it can be used for both sentence-to-clip and clip-to-sentence retrieval. The clip is enriched with the context, whether used in the gallery set (in sentence-to-clip) or in the query (in clip-to-sentence). When performing sentence-to-clip retrieval, our query consists of only one sentence as usually done in other approaches, and is thus comparable to these. 
During inference, the gallery of clips is always given, and thus all approaches have access to the same information.
\vspace*{-10pt}
\subsection{Multi-modal Context}
\vspace*{-6pt}
\label{sec:method:modality}
In previous sections (Sec.~\ref{subsec:method:definition}--\ref{subsec:method:losses}), we motivated our approach by focusing on local clip context---i.e. context in the visual modality.
However, ConTra could similarly be applied to the local context of the text modality.
As an example, given steps of a recipe, these can be utilised to build a Text Context (TC), such that:
\begin{equation}
TC_m(t_{ij}) = (t_{i(j-m)}, \cdots, t_{ij}, \cdots, t_{i(j+m)})
\end{equation}

To give an example for clip-to-sentence retrieval, a single clip is used as the query, but the gallery is constructed of captions that have had their representation enriched via sentences of neighbouring clips (e.g. ``Add the mince to the pan'' has attended to ``Take the beef mince out of its wrapper'' and ``Fry until the mince is browned''). This contextual text knowledge could come from video narrations or steps in a recipe. 


 We also assess the utilisation of context in both modalities.
This setup assumes access to local context in both clip and sentence. 
$L_{NEI}$ is thus applied to both neighbouring clip contexts and text contexts, using one negative for each case.
The architecture for both $f_{cc}$ and $f_{tc}$ are identical, but are learned as two separate embedding functions with unique weights and positional embeddings\footnote{We experimented with sharing these embeddings but similar to previous approaches~\cite{DBLP:conf/icml/JaegleGBVZC21}, this performed worse, see supplementary.}.

\section{Results}

We first present our experimental settings and the choice of untrimmed datasets in Sec.~\ref{sec:exp-settiing}. 
We then focus on clip context results including comparison with state-of-the-art (SOTA) methods in Sec.~\ref{sec:exp-clip-context} before exploring text context and context in both modalities in 
Sec.~\ref{sec:exp-modality-context}.
Finally, we discuss limitations and avenues for future work.

\vspace*{-10pt}
\subsection{Experimental Settings}
\label{sec:exp-settiing}
\vspace*{-6pt}
\noindent
\textbf{Datasets.}\quad Video datasets commonly used for cross-modal retrieval can be split into two groups: trimmed and untrimmed.
In trimmed datasets, such as MSRVTT~\cite{DBLP:conf/cvpr/XuMYR16}, MSVD~\cite{DBLP:conf/acl/ChenD11} and VATEX~\cite{DBLP:conf/iccv/WangWCLWW19}, the full video is considered as a single clip and thus no context can be utilised. 
In Table~\ref{tab:datasets}, we compare the untrimmed datasets for their size and the number of clips per video.
Datasets with 1-2 clips per video on average limit the opportunity to explore long or local temporal context.
While we include QuerYD~\cite{DBLP:conf/icassp/OncescuHLZA21} 
in the table, this dataset does not allow for context to be explored as clips from the same video are split between the train and test sets. 
We choose to evaluate our method on three untrimmed datasets, whose average number of clips/video is greater than $3$.
We describe the notion of context in each:

YouCook2~\cite{DBLP:conf/aaai/ZhouXC18} contains YouTube cooking videos. On average, training videos contain 7.75 clips, each associated with a sentence.
The dataset has been evaluated for clip-sentence retrieval~\cite{DBLP:journals/corr/abs-2105-13033,DBLP:conf/naacl/TangLB21,DBLP:conf/cvpr/MiechASLSZ20,DBLP:journals/corr/abs-2104-00285} as well as video-paragraph retrieval~\cite{DBLP:conf/nips/GingZPB20}. We focus on clip-sentence retrieval, utilising the local context, which represents previous/follow-up steps in a recipe. 
Given YouCook2's popularity, we use it for all ablation experiments.

ActivityNet Captions~\cite{DBLP:conf/iccv/KrishnaHRFN17} consists of 
annotated YouTube videos from ActivityNet~\cite{DBLP:conf/cvpr/HeilbronEGN15}.
The dataset has only been evaluated for video-paragraph retrieval~\cite{DBLP:conf/nips/GingZPB20,DBLP:conf/eccv/ZhangHS18,DBLP:journals/corr/abs-2102-06183} where all clips in the same video are concatenated, and all corresponding captions are also concatenated to form the paragraph.
Instead, we consider the \emph{val\_1} split, and introduce an \textbf{ActivityNet Clip-Sentence (CS)} variant using all the individual clips and their corresponding captions/sentences. 
We emphasise that this evaluation \emph{cannot} be compared to published results on video-paragraph retrieval and instead evaluate two methods to act as baselines for comparison.

\begin{table}[t]
\begin{center}
\begin{tabular}{l|cc|cc|cc}
&  \multicolumn{2}{c|}{\textbf{\#clips}}& \multicolumn{2}{c|}{\textbf{\#videos}}& \multicolumn{2}{c}{\textbf{\#clips per video}}  \\
\cline{2-7}
\multicolumn{1}{l}{Datasets} &\multicolumn{1}{|c}{Train} & \multicolumn{1}{c}{Test}&\multicolumn{1}{|c}{Train} & \multicolumn{1}{c}{Test}&\multicolumn{1}{|c}{Train} & \multicolumn{1}{c}{Test} \\\hline
Charades-STA ~\cite{DBLP:conf/iccv/GaoSYN17} & 5,657 & 1,596 &5338 & 1334 & 1.60 & 1.20    \\
DiDeMo~\cite{DBLP:conf/emnlp/HendricksWSSDR18} & 21,648 & 2,650 &8511 & 1037 & 2.21 & 2.21   \\
QuerYD*~\cite{DBLP:conf/icassp/OncescuHLZA21} & 9,118 & 1,956 & 1,283 & 794 &\multicolumn{2}{c}{8.3}\\ \hline
YouCook2~\cite{DBLP:conf/aaai/ZhouXC18} & 10,337 & 3,492 &1,333 & 457 & 7.75 & 7.64 \\
ActivityNet CS~\cite{DBLP:conf/iccv/KrishnaHRFN17} & 37,421 & 17505 &10,009 & 4,917 & 3.74 & 3.56    \\
EPIC-KITCHENS-100~\cite{DBLP:journals/corr/abs-2006-13256} & 67,217 & 9,668 &495 & 138 & 135.79 & 70.06  \\\hline
\end{tabular}

\caption{Comparing untrimmed video datasets by size and number of clips per video. *QuerYD videos are split across train and test so we report overall clips/video.}
\vspace*{-35pt}
\label{tab:datasets}
\end{center}
\end{table}

EPIC-KITCHENS-100~\cite{DBLP:journals/corr/abs-2006-13256} offers a unique opportunity to explore context in significantly longer untrimmed videos. On average, there are 135.8 clips per video of kitchen-based actions, shot from an egocentric perspective. 
We use the train/test splits for the multi-instance retrieval benchmark, but evaluate on the single-instance retrieval task using the set of unique narrations.

\noindent
\textbf{Evaluation Metrics.}\quad We report the retrieval performance, for both clip-to-sentence and sentence-to-clip tasks, using the two standard metrics of: Recall at $K=\{1,5,10\}$ (R@K) and median rank (MR). We also report the sum of cross-modal R@K as RSum to demonstrate overall performance. Where figures are plotted, tables of exact results are given in supplementary.

\noindent
\textbf{Visual Features.}\quad To be comparable to prior work, we use the same features as recent methods per dataset. For YouCook2, we use the S3D backbone provided by~\cite{DBLP:conf/cvpr/MiechASLSZ20} pre-trained on~\cite{DBLP:conf/iccv/MiechZATLS19} , extracting $1024$-$d$ features. 
We uniformly sample 32 frames from each clip with a 224x224 resolution. 
For ActivityNet CS, we use frame features provided by~\cite{DBLP:conf/eccv/ZhangHS18}. These frame features are combined into clip features using a single transformer, trained with shared weights across clips, as proposed in~\cite{DBLP:conf/nips/GingZPB20}, obtaining $384$-$d$ features.
Note that this transformer is trained for clip-sentence alignment, using the code from~\cite{DBLP:conf/nips/GingZPB20},
without the global context and contextual transformer.
For EPIC-KITCHENS-100, we use the publicly available $3072$-$d$ features from~\cite{kazakos2019TBN}. 

\noindent
\textbf{Text Features.}\quad For YouCook2 and EPIC-KITCHENS-100, we take a maximum of 16 words without removing stopwords from each sentence and we extract $2048$-$d$ feature vectors using the text branch in~\cite{DBLP:conf/cvpr/MiechASLSZ20} pre-trained on~\cite{DBLP:conf/iccv/MiechZATLS19}.
This consists of a linear layer with a ReLU activation applied independently to each word embedding followed by max pooling and a randomly initialised linear layer to reduce dimensionality.  We fine-tune the text branch layer, to accommodate missing vocabulary\footnote{We add 174 and 104 missing words from the model in~\cite{DBLP:conf/iccv/MiechZATLS19} for YouCook2 and EPIC-KITCHENS-100 respectively.}. 
For ActivityNet CS, we feed the sentences into a pretrained BERT-Base
Uncased model~\cite{DBLP:conf/naacl/DevlinCLT19} and use the per-token outputs of the last 2 layers to train a sentence transformer, as in~\cite{DBLP:conf/nips/GingZPB20}, and obtain $384$-$d$ text features.

\noindent
\textbf{Architecture Details.}\quad 
The number of layers and heads in the ConTra encoder differs depending on the dataset size. 
For the small-scaled YouCook2, we use $1$ layer and $R=2$ heads to avoid overfitting.
For the larger two datasets we use $2$ layers and $R=8$ heads.
The inner dimension of the transformers is $2048$-$d$. The learnt positional encoding matches the feature dimension: $512$-$d$ for YouCook2 and EPIC-KITCHENS-100, and $384$-$d$ for ActivityNet CS, initialised from $\mathcal{N}(0,0.001)$. 
Due to the small dimension of the features of ActivityNet CS, we remove the linear projection $P$ from our architecture.
We apply dropout of $0.3$ at $h_j$ and~$h_j'$. 

\noindent
\textbf{Implementation Details.}\quad 
We use the Adam optimizer with a starting learning rate of $1 \times 10^{-4}$ and decrease it linearly  after a warmup period of $1300$ iterations. 
The size of the batch is fixed to $512$. The temperature $\tau$ in Eq.~\ref{eqn:NCE} and~\ref{eqn:Neighbouring} is set to $0.07$ as in~\cite{DBLP:conf/cvpr/He0WXG20,DBLP:journals/corr/abs-2003-04298,DBLP:conf/cvpr/WuXYL18}, and the dimension of the common embedding space $\Omega$ is set to $512$ for YouCook2 and EPIC-KITCHENS-100, and $384$ for ActivityNet CS.
We ablate these values in supplementary.
If the clip does not have sufficient temporal context (i.e. is at the start/end of the video), we pad the input by duplicating the first/last clip to obtain a fixed-length context.  
\emph{Our code is available at \url {https://github.com/adrianofragomeni/ConTra}}
\vspace*{-10pt}
\subsection{Clip Context Results}
\label{sec:exp-clip-context}
\begin{wrapfigure}[14]{r}{.5\linewidth}
\centering
\vspace*{-17pt}
    \includegraphics[width=\linewidth]{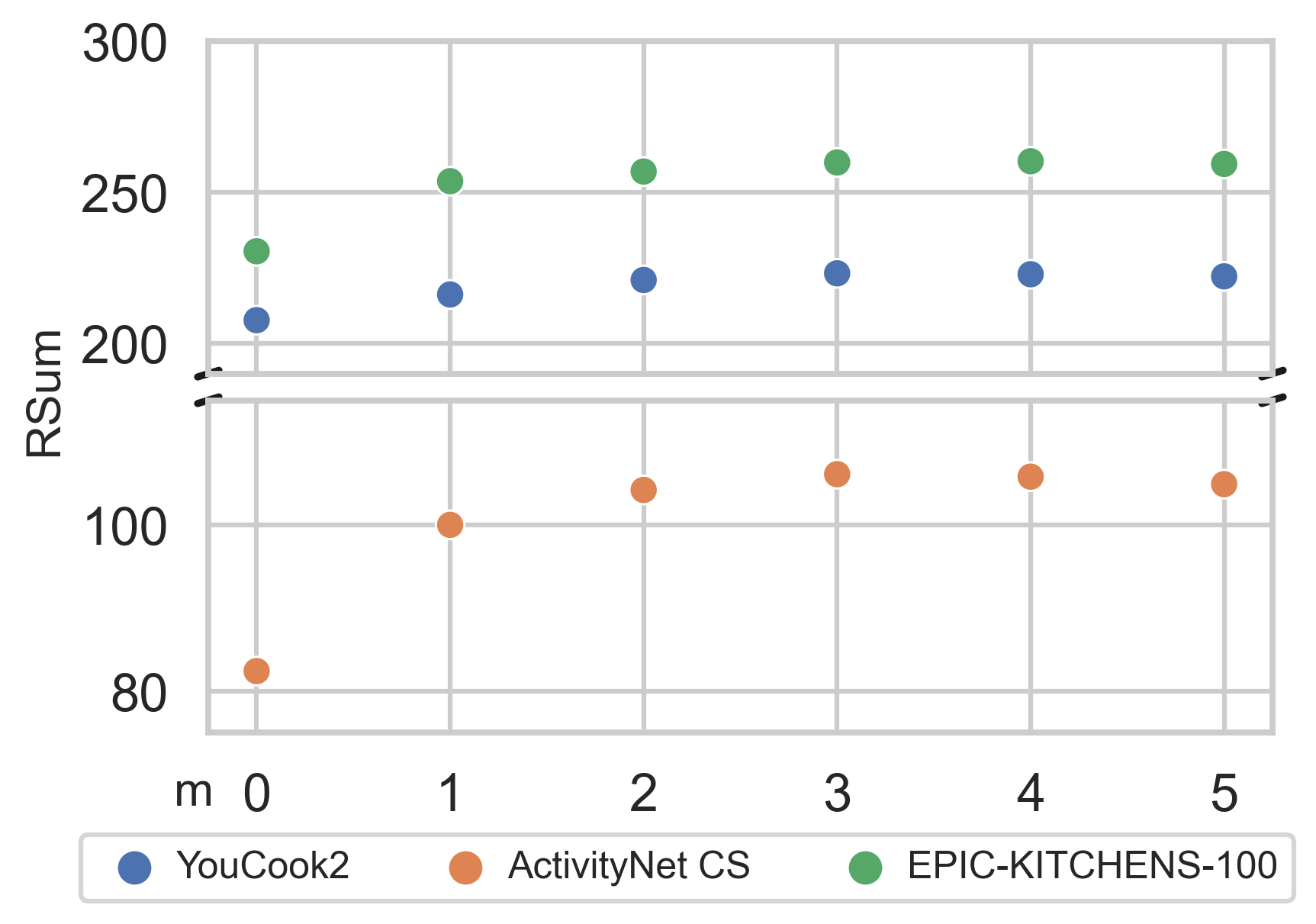}
    \vspace*{-20pt}
    \caption{Analysis of clip context (CC) with differing $m$ across YouCook2, ActivityNet CS and EPIC-KITCHENS-100.}
    \label{fig:fig_contextvideo}
\end{wrapfigure}

\begin{figure*}[t]
    \centering
    \includegraphics[width=\linewidth]{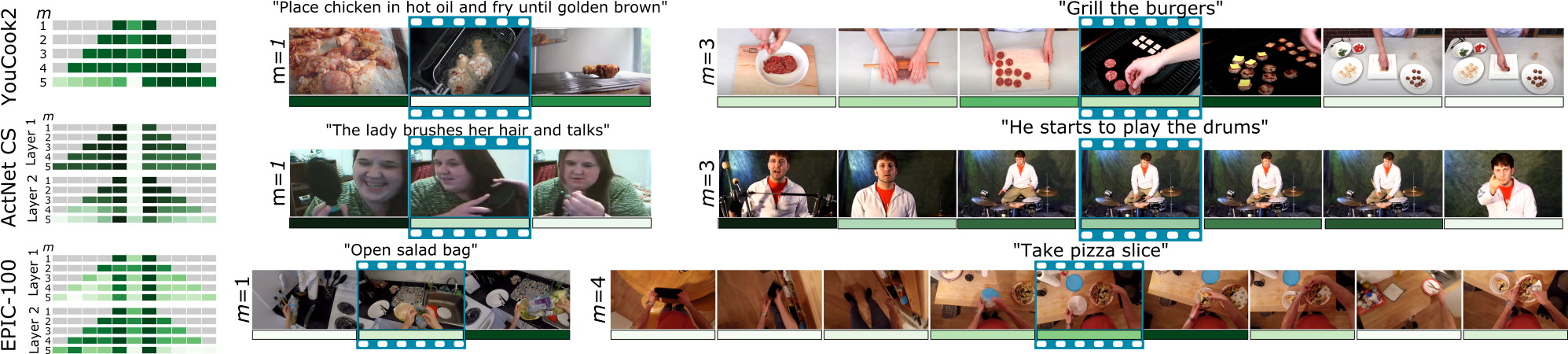}
    \caption{Left: average attention weights over videos as $m$ changes, per dataset and layer. Right: qualitative examples with clip attention.}
    \vspace*{-12pt}
    \label{fig:attention-weights}
\end{figure*}

\vspace*{-6pt}
\paragraph{Context Length Analysis.}\quad
We analyse the effect of clip context length by varying $m$ from its no-context baseline, $m=0$, to $m=5$.  Results are presented in Fig.~\ref{fig:fig_contextvideo}. In all three datasets, the largest improvement is obtained comparing $m=0$, i.e. no context, to $m=1$, i.e. introducing the smallest context, where the RSum increases by $8.5$, $17.6$ and $23.2$ for Youcook2, ActivityNet CS, and EPIC-KITCHENS-100, respectively. This highlights that neighbouring clips are able to improve the retrieval performance. Moreover, every $m > 0$ outperforms $m=0$ on all datasets.
We obtain the best performance on YouCook2 at $m = 3$. 
ActivityNet CS also obtained best performance when using $m=3$, and EPIC-KITCHENS-100 when $m=4$. 
Although ConTra introduces a new hyperparameter, $m$, Fig.~\ref{fig:fig_contextvideo} shows that RSum saturates when $m \geq 3$ across all datasets. Using a larger context does not further improve the performance.
We show the attention weights learned by the multi-headed attention layers in Fig.~\ref{fig:attention-weights} averaged over all videos, per layer (left), and for specific examples~(right).
YouCook2 focuses more on the later clips due to the recipes being more recognisable when ingredients are brought together.
The attention weights for EPIC-KITCHENS-100 and ActivityNet CS are higher for earlier clips, with ActivityNet's first layer and EPIC-KITCHENS-100's second layer attending to past clips.
EPIC-KITCHENS-100 specifically has higher attention weights on directly neighbouring clips.
From the examples in Fig.~\ref{fig:attention-weights} (right), ConTra uses local context to discriminate objects which may be occluded, such as chicken in YouCook2, the contents of the salad bag in EPIC-KITCHENS-100, or the brush in ActivityNet CS.

\begin{wrapfigure}[18]{r}{.5\linewidth}
    \vspace*{-18pt}
    \includegraphics[width=.5\textwidth]{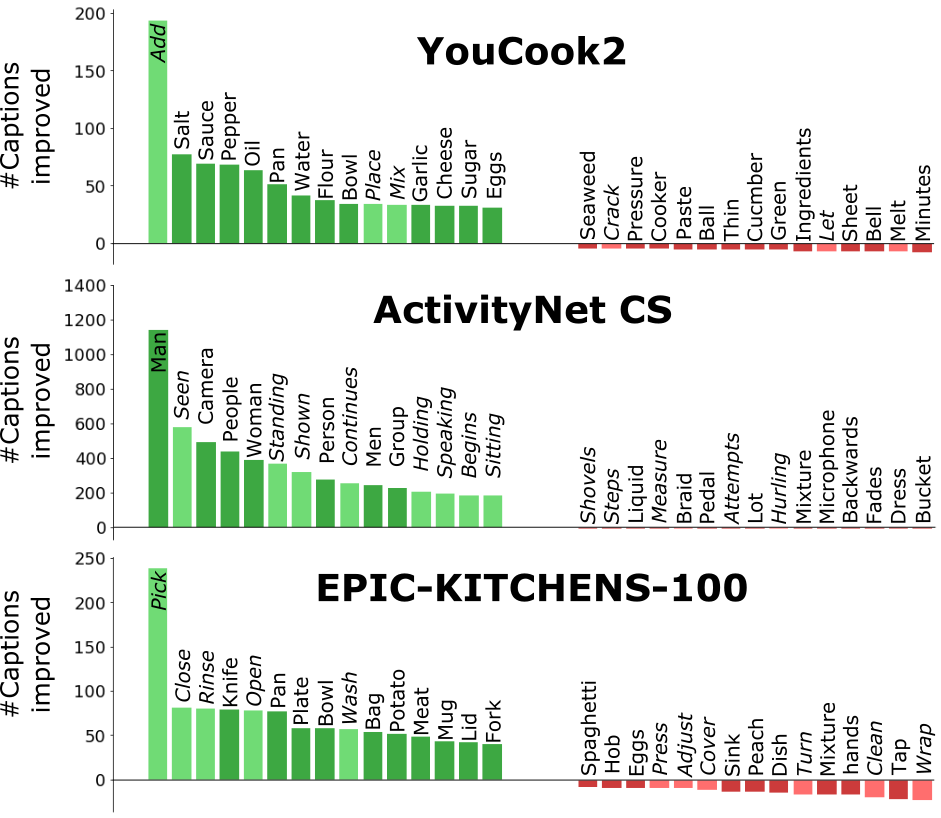}
    \vspace*{-12pt}
    \caption{15 most improved/hindered words per dataset when context is used. Verbs are lighter/italicised. Best viewed in colour.}
    \vspace*{-15pt}
    \label{fig:rank_changes}
\end{wrapfigure}

In Fig.~\ref{fig:rank_changes}, we analyse how individual words are affected by using context. On a word-by-word basis we find all captions that contain a given word and
count the number of times the rank of those captions improved/worsened after adding context. E.g., the word `pan’
in YouCook2 is present in 373 captions, in which 182 captions improve their rank with context while 126 captions worsen their rank resulting in a delta of 56.
In YouCook2, `salt', `sauce', and `oil' all see a large improvement when using context, likely due to easily being obscured by their containers.
In comparison, for EPIC-KITCHENS-100, verbs benefit the most from context---these actions tend to be very short so surrounding context can help discriminate them.

Overall, these results showcase that using local Clip Context (CC) enhances the clip embedding representation and consistently results in a boost in performance. 

\paragraph{Comparison with State of the Art.}\quad
The most commonly used untrimmed dataset in prior work is YouCook2.
For fair comparison, we split the SOTA methods on this dataset into blocks according to the pre-training and fine-tuning datasets: (i)~training only on YouCook2, (ii)~training only on other large-scale datasets, (iii)~pre-training on large-scale datasets then fine-tuning on YouCook2; note that this is where ConTra lies and (iv)~additionally, pre-training with proxy tasks on large-scale datasets.
Table~\ref{tab:comparisonY} compares ConTra with the SOTA on YouCook2.

Overall, ConTra outperforms all directly-comparable SOTA works~\cite{DBLP:conf/iccv/MiechZATLS19,DBLP:journals/corr/abs-2105-13033,DBLP:conf/nips/GingZPB20}. ConTra outperforms COOT~\cite{DBLP:conf/nips/GingZPB20} that trains for video-paragraph retrieval on the full video. 
Distinct from COOT~\cite{DBLP:conf/nips/GingZPB20}, we only use clip context, i.e. single sentences in training and inference, and local context.

\begin{table}[t!]
\vspace{-1mm}
\begin{center}

\begin{tabular}{P{20pt}p{95pt}P{15pt}P{15pt}|P{25pt}P{25pt}P{30pt}P{25pt}|P{25pt}}

&\multicolumn{1}{l}{Method}&
\multicolumn{1}{c}{PX}&\multicolumn{1}{c|}{FT}&\multicolumn{1}{c}{R@1} &\multicolumn{1}{c}{R@5}&\multicolumn{1}{c}{R@10} &\multicolumn{1}{c|}{MR} 
&\multicolumn{1}{c}{RSum} \\\hline
\multirow{2}{*}{(i)} &HGLMM~\cite{DBLP:conf/cvpr/KleinLSW15} & $\times$ & $\checkmark$ & 4.6 & 14.3 & 21.6 & 75 & 40.5  \\
&UniVL (FT-joint)~\cite{DBLP:journals/corr/abs-2002-06353} &   $\times$ &$\checkmark$& 7.7 & 23.9 & 34.7 & 21 & 66.3 \\\hline

\multirow{5}{*}{(ii)} &ActBert~\cite{DBLP:conf/cvpr/ZhuY20a} &  \checkmark &$\times$ & 9.6 & 26.7 & 38.0 & 19 & 74.3  \\
&MMV FAC~\cite{DBLP:conf/nips/AlayracRSARFSDZ20} &  $\times$ &$\times$ &  11.7 & 33.4 & 45.4 & 13 & 90.5  \\
&VATT-MBS~\cite{DBLP:journals/corr/abs-2104-11178} & $\times$& $\times$ & - & - & 45.5 & 13 & -  \\
&MCN~\cite{DBLP:journals/corr/abs-2104-12671} & $\times$& $\times$ & 18.1 & 35.5 & 45.2 & - & 98.8  \\
&MIL-NCE~\cite{DBLP:conf/cvpr/MiechASLSZ20}& $\times$& $\times$ & 15.1 & 38.0 & 51.2 & 10 & 104.3   \\\hline

\multirow{5}{*}{(iii)} &HowTo100M~\cite{DBLP:conf/iccv/MiechZATLS19} & $\times$ & $\checkmark$ & 8.2 & 24.5 & 35.3 & 24 & 68.0   \\
&GRU+SSA~\cite{DBLP:journals/corr/abs-2105-13033} & $\times$ & $\checkmark$& 10.9 & 28.4 & - & - & - \\
&COOT~\cite{DBLP:conf/nips/GingZPB20}& $\times$ & $\checkmark$ & 16.7 & 40.2 & 52.3 & 9 & 109.2  \\
&MIL-NCE (from~\cite{DBLP:journals/corr/abs-2104-00285}) & $\times$ & $\checkmark$& 15.8 & 40.3 & 54.1 & 8 & 110.2   \\
&\bfseries{ConTra (ours)} & $\times$ & $\checkmark$& 16.7 & 42.1 & 55.2 & 8 & 114.0  \\\hline \hline

\multirow{5}{*}{(iv)} &CUPID~\cite{DBLP:journals/corr/abs-2104-00285} & $\checkmark$& \checkmark & 17.7 & 43.2 & 57.1 & 7 &  117.9 \\
&DeCEMBERT~\cite{DBLP:conf/naacl/TangLB21} & $\checkmark$& \checkmark & 17.0 & 43.8 & 59.8 & 9 & 120.6  \\
&UniVL (FT-joint)~\cite{DBLP:journals/corr/abs-2002-06353} &  $\checkmark$& \checkmark & 22.2 & 52.2 & 66.2 & 5 & 140.6 \\
&VLM~\cite{DBLP:conf/acl/XuGHAAFMZ21} &  $\checkmark$& \checkmark & 27.0 & 56.9 & 69.4 & 4 & 153.3 \\
&VideoCLIP~\cite{xu-etal-2021-videoclip} &  $\checkmark$& \checkmark & 32.2 & 62.6 & 75.0 & - & 169.8 \\\hline
\end{tabular}
\caption{Sentence-to-Clip comparison with SOTA on YouCook2 test set. PX: pre-train end-to-end with proxy tasks. FT: fine-Tuning on YouCook2. $-$: unreported results.}
\vspace*{-18pt}
\label{tab:comparisonY}
\end{center}
\end{table}

\begin{table}[t!]
\vspace{-2mm}
\begin{center}

\begin{tabular}{P{20pt}p{75pt}P{10pt}|P{20pt}P{20pt}P{20pt}P{20pt}|P{20pt}P{20pt}P{20pt}P{20pt}|P{20pt}}

&&&\multicolumn{4}{c|}{\textbf{Sentence-to-Clip}} & \multicolumn{4}{c|}{\textbf{Clip-to-Sentence}} & \\
\cline{4-11}

&\multicolumn{1}{l}{Method}&\multicolumn{1}{c|}{FT}&\multicolumn{1}{c}{\smaller{R@1}} &\multicolumn{1}{c}{\smaller{R@5}}&\multicolumn{1}{c}{\smaller{R@10}} &\multicolumn{1}{c|}{\smaller{MR}} 
&\multicolumn{1}{c}{\smaller{R@1}} &\multicolumn{1}{c}{\smaller{R@5}}&\multicolumn{1}{c}{\smaller{R@10}} &\multicolumn{1}{c|}{\smaller{MR}}&
\multicolumn{1}{c}{\smaller{RSum}} \\\hline

(ii)&MIL-NCE~\cite{DBLP:conf/cvpr/MiechASLSZ20}$^{*}$ &$\times$ & 2.4 & 6.8 & 10.0 & 460 & 2.1 & 6.2 & 9.2 & 543 & 36.7  \\\hline

\multirow{4}{*}{(iii)}&HowTo100M~\cite{DBLP:conf/iccv/MiechZATLS19}$^{*}$ &$\checkmark$  & 3.8 & 12.5 & 18.9 & 68 & 3.6 & 11.4 & 17.3 & 78 & 67.5  \\
&COOT~\cite{DBLP:conf/nips/GingZPB20} $-g^{*}$ &$\checkmark$ &3.7 & 11.9 & 18.6 & 67 & 3.7 & 12.0 & 18.8 & 64 & 68.7\\
&\bfseries{ConTra (ours)}&$\checkmark^\dagger$ &5.9 & 18.4 & 27.6 & 38 & 6.4 & 19.3 & 28.5 & 37 & 106.1\\
&COOT~\cite{DBLP:conf/nips/GingZPB20}$^{*}$&$\checkmark$ & 6.2 & 18.8 & 28.4 & 33 & 6.3 & 19.0 & 28.4 & 32 & 107.0\\\hline
\end{tabular}
\caption{Comparison on ActivityNet CS for Clip-Sentence Retrieval. $^{*}$Our reproduced results. $^\dagger$:  Transformer fine-tuned first---clip/sentence features match those from COOT~\cite{DBLP:conf/nips/GingZPB20} $-g$.}
\vspace*{-24pt}
\label{tab:comparisonActNet}
\end{center}
\end{table}

\begin{table}[t!]
\vspace{-1mm}
\begin{center}

\begin{tabular}{P{20pt}p{75pt}P{10pt}|P{20pt}P{20pt}P{20pt}P{20pt}|P{20pt}P{20pt}P{20pt}P{20pt}|P{20pt}}

&&&\multicolumn{4}{c|}{\textbf{Sentence-to-Clip}} & \multicolumn{4}{c|}{\textbf{Clip-to-Sentence}} & \\
\cline{4-11}

&\multicolumn{1}{l}{Method}&\multicolumn{1}{c|}{FT}&\multicolumn{1}{c}{\smaller{R@1}} &\multicolumn{1}{c}{\smaller{R@5}}&\multicolumn{1}{c}{\smaller{R@10}} &\multicolumn{1}{c|}{\smaller{MR}} 
&\multicolumn{1}{c}{\smaller{R@1}} &\multicolumn{1}{c}{\smaller{R@5}}&\multicolumn{1}{c}{\smaller{R@10}} &\multicolumn{1}{c|}{\smaller{MR}}&
\multicolumn{1}{c}{\smaller{RSum}} \\\hline

(ii)&MIL-NCE~\cite{DBLP:conf/cvpr/MiechASLSZ20}$^{*}$ &$\times$ & 3.2 & 10.4 & 15.4 & 188 & 2.1 & 7.8 & 12.3 & 194 & 51.2  \\\hline

\multirow{2}{*}{(iii)}&JPoSE~\cite{DBLP:conf/iccv/WrayCLD19}* &$\checkmark$& 2.5 & 7.5 & 11.6 & 13 & 4.4 & 17.4 & 27.2 & 17 & 70.6  \\
&\bfseries{ConTra (ours)} & $\checkmark$& 22.2 & 43.4 & 53.4 & 9 & 28.2 & 52.0 & 61.1 & 5 & 260.3\\\hline
\end{tabular}
\caption{Comparison with baseline on EPIC-KITCHENS-100. $^{*}$Our reproduced results.}
\vspace*{-32pt}
\label{tab:comparisonEPIC}
\end{center}
\end{table}

The last block of Table~\ref{tab:comparisonY} includes works that are not directly comparable to ConTra, as these models are pre-trained \emph{end-to-end} on HowTo100M with additional proxy tasks, e.g. masked language modelling, whereas ConTra is initialised randomly. 
Although DeCEMBERT~\cite{DBLP:conf/naacl/TangLB21} is not directly comparable, ConTra is less complex with $9.5M$ parameters compared to DeCEMBERT's $115.0M$ and our results are only marginally lower.

To the best of our knowledge, no prior work has evaluated on ActivityNet CS for clip-sentence retrieval. For comparison, we evaluate~\cite{DBLP:conf/cvpr/MiechASLSZ20,DBLP:conf/iccv/MiechZATLS19} on ActivityNet CS for clip-sentence retrieval using public code.
We also run the code from COOT~\cite{DBLP:conf/nips/GingZPB20}, trained on video-paragraph retrieval, and obtain their results on clip-sentence retrieval. 
We then replace the text input with sentence-level representations, and remove their global alignment to produce the COOT$-g$ baseline reported above.
Table~\ref{tab:comparisonActNet} shows that ConTra outperforms MIL-NCE~\cite{DBLP:conf/cvpr/MiechASLSZ20}, HowTo100M~\cite{DBLP:conf/iccv/MiechZATLS19} and COOT$-g$ by a considerable margin. 
Our RSum is only marginally lower than COOT, where global context is considered for both modalities during training. 
\begin{wrapfigure}[13]{r}{.5\linewidth}
    \centering
    \vspace*{-20pt}
    \includegraphics[width=\linewidth]{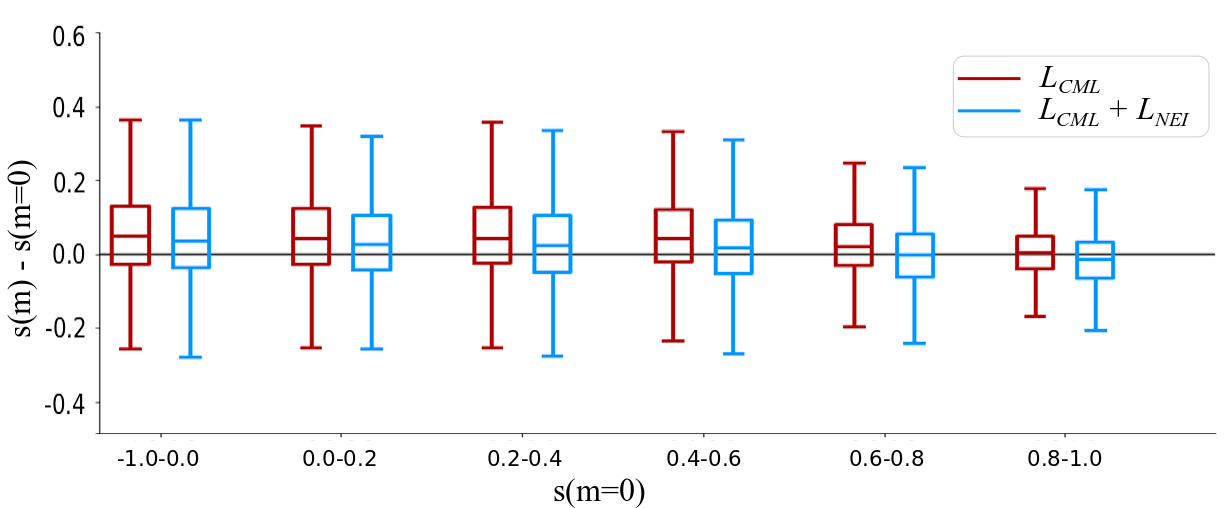}
    \caption{Comparison between similarities to neighbouring clips, $s(f_{cc}(j+1),f_s(j))$, with and without using $L_{NEI}$. Without $L_{NEI}$ ConTra gives higher similarities to neighbouring clips.}
    \label{fig:neighbouring_loss_comp}
    \vspace*{-12pt}
\end{wrapfigure}

Note that methods that train for global context cannot be used for datasets with hundreds of clips per video, like EPIC-KITCHENS-100.
In Table~\ref{tab:comparisonEPIC}, we compare ConTra to JPoSE~\cite{DBLP:conf/iccv/WrayCLD19} and  our reproduced results of MIL-NCE~\cite{DBLP:conf/cvpr/MiechASLSZ20} on EPIC-KITCHENS-100, outperforming on all metrics by a large margin. We cannot train or evaluate COOT on EPIC-KITCHENS-100 which has 136 clips per video on average. Additionally, clips in EPIC-KITCHENS-100 are significantly shorter, increasing the benefits of attending to local context.

\paragraph{Ablation studies.}\quad
We ablate ConTra on YouCook2. Ablations on the other two datasets are in supplementary.

\noindent
\textbf{Loss Function.}\quad In Table~\ref{tab:loss}, we first test the formulation of the two losses $L_{NEI}$ and $L_{CML}$ individually, using the NCE loss~\cite{DBLP:conf/acl/XuGHAAFMZ21,DBLP:journals/corr/abs-2104-11178,DBLP:journals/corr/abs-2104-12671,Liu2021HiTHT}.
%
\begin{table}[t]
\begin{center}
\begin{tabular}{l|P{25pt}P{25pt}P{25pt}P{25pt}|P{25pt}P{25pt}P{25pt}P{25pt}|P{25pt}}
&\multicolumn{4}{c|}{\textbf{Sentence-to-Clip}} & \multicolumn{4}{c|}{\textbf{Clip-to-Sentence}} & \\
\cline{2-9}
\multicolumn{1}{l|}{Loss}
&\multicolumn{1}{c}{\smaller{R@1}} &\multicolumn{1}{c}{\smaller{R@5}}&\multicolumn{1}{c}{\smaller{R@10}} &\multicolumn{1}{c|}{\smaller{MR}} 
&\multicolumn{1}{c}{\smaller{R@1}} &\multicolumn{1}{c}{\smaller{R@5}}&\multicolumn{1}{c}{\smaller{R@10}} &\multicolumn{1}{c|}{\smaller{MR}}&
\multicolumn{1}{c}{\smaller{RSum}} \\\hline
$L_{NEI}$& 6.4 & 18.3 & 27.3 & 39 & 4.3 & 15.5 & 23.6 & 43 & 95.4\\
$L_{CML}$ & 15.7 & 39.9 & 53.4 & 9 & 14.5 & 38.5 & 52.3 & \textbf{9} & 214.3\\
$L_{CML}$+$L_{HardMining}$ & 15.7 & 39.8 & 53.5 & 9 & 14.2 & 39.0 & 51.9 & 10 & 214.1\\
$L_{CML}$+$L_{NEI}$ & 16.2 & 41.4 & 54.1 & 9 & \textbf{14.8} & 39.3 & 52.6 & \textbf{9} & 218.4\\
$L_{CML}$+$L_{NEI}$+$L_{UNI}$ &\textbf{16.7} & \textbf{42.1} & \textbf{55.2} & \textbf{8} & \textbf{14.8} & \textbf{40.5} & \textbf{53.9} & \textbf{9} & \textbf{223.2}\\ \hline
\end{tabular}
\caption{Ablation of loss function terms: Neighbouring Loss ($L_{NEI}$), Cross Modal Loss ($L_{CML}$), Hard Triplet Mining ($L_{HardMining}$), and Uniformity Loss ($L_{UNI}$).}
\label{tab:loss}
\vspace*{-34pt}
\end{center}
\end{table}
On its own, $L_{NEI}$ learns with limited variety of only neighbouring clips as negatives.  

Then, we compare $L_{NEI}$ to the standard hard mining approach proposed in~\cite{DBLP:conf/bmvc/FaghriFKF18}. $L_{NEI}$ consistently outperforms hard mining. 
Our proposed loss $L$, with its 3 terms, performs the best, improving RSum by~$4.8$ when adding the uniformity loss $L_{UNI}$ which allows preserving maximal information and so obtains better embeddings.

We further demonstrate the benefits of $L_{NEI}$ in Fig.~\ref{fig:neighbouring_loss_comp}.
We bin the neighbouring clips $j \pm 1$ based on their similarity to the sentence $j$ along the x-axis.
We then calculate the difference between this similarity with and without context, and provide the average and extent of these differences in a box plot over all datasets.
When this difference, on the y-axis, is $> 0$, the context transformer would have increased the similarity between the neighbouring clip and the sentence.
Without $L_{NEI}$, the similarity is increased further, particularly for clips and sentences with low cosine similarity, depicted on the x-axis.

\begin{wraptable}[9]{r}{.6\linewidth}
\begin{center}
\resizebox{1\linewidth}{!}{%
\begin{tabular}{c|cccc|cccc|c}

&\multicolumn{4}{c|}{\textbf{Sentence-to-Clip}} & \multicolumn{4}{c|}{\textbf{Clip-to-Sentence}} & \\
\cline{2-9}

\multicolumn{1}{c|}{\#Negatives} &\multicolumn{1}{c}{R@1} &\multicolumn{1}{c}{R@5}&\multicolumn{1}{c}{R@10} &\multicolumn{1}{c|}{MR} 
&\multicolumn{1}{c}{R@1} &\multicolumn{1}{c}{R@5}&\multicolumn{1}{c}{R@10} &\multicolumn{1}{c|}{MR}&
\multicolumn{1}{c}{RSum} \\\hline
1& 16.7 & 42.1 & 55.2 & \textbf{8} & 14.8 & 40.5 & \textbf{53.9} & \textbf{9} & 223.2\\
2 & 16.9 & 42.2 & \textbf{55.7} & \textbf{8} & \textbf{15.6} & 40.5 & \textbf{53.9} & \textbf{9} & \textbf{224.8}\\
3 & \textbf{17.3} & \textbf{42.4} & 55.6 & \textbf{8} & 14.9 & \textbf{40.7} & \textbf{53.9} & \textbf{9} & \textbf{224.8}\\\hline
\end{tabular}}
\caption{Analysis of the performance varying the number of negative in $L_{NEI}$.}
\label{tab:negatives_NEI}
\end{center}
\end{wraptable}

\noindent
\textbf{Number of negatives neighbouring clips.}\quad Table~\ref{tab:negatives_NEI} shows how the performance changes when we consider more than one negative for our neighbouring loss $L_{NEI}$. Increasing the negatives from $1$ to $2$ improves the retrieval results marginally. RSum remains the same when increasing the number of negatives further to $3$. We keep the number of negatives equal to $1$ in all the experiments.

\noindent
\textbf{Aggregate context.}\quad 
As explained in Sec.~\ref{subsec:method:contra}, we select the middle output of the transformer encoder as our clip embedding. In order to justify this design choice, we compare to other aggregation approaches. These are of two types based on where local context is aggregated, i.e the visual features $h_{j}$ or the outputs of the clip transformer encoder $f_{cc}$. Moreover we experimented with two aggregation techniques, Maximum and Average. 
\begin{wraptable}[7]{r}{.6\linewidth}
\begin{center}
\vspace*{-32pt}
\resizebox{\linewidth}{!}{%
\begin{tabular}{lc|cccc|cccc|c}
&&\multicolumn{4}{c|}{\textbf{Sentence-to-Clip}} & \multicolumn{4}{c|}{\textbf{Clip-to-Sentence}} & \\
\cline{3-10}
\multicolumn{1}{c}{$h_{j}$} &\multicolumn{1}{c|}{$f_{cc}$} &\multicolumn{1}{c}{R@1} &\multicolumn{1}{c}{R@5}&\multicolumn{1}{c}{R@10} &\multicolumn{1}{c|}{MR} 
&\multicolumn{1}{c}{R@1} &\multicolumn{1}{c}{R@5}&\multicolumn{1}{c}{R@10} &\multicolumn{1}{c|}{MR}&
\multicolumn{1}{c}{RSum} \\\hline
Avg&-  &4.9 & 16.3 & 26.3 & 40 & 4.3 & 16.0 & 25.8 & 41 & 93.6\\
Max&-  &3.1 & 13.1 & 22.0 & 46 & 3.4 & 13.4 & 21.7 & 49 & 76.7\\ \hline
-&Avg  &15.6 & 40.1 & 53.9 & 9 & 14.0 & 38.6 & 51.6 & 10 & 213.8\\
-&Max  &\textbf{16.9} & 41.7 & 54.8 & \textbf{8} & 14.7 & 40.1 & 53.7 & \textbf{9} & 221.9\\
-&Mid  &16.7 & \textbf{42.1} & \textbf{55.2} & \textbf{8} & \textbf{14.8} & \textbf{40.5} & \textbf{53.9} & \textbf{9} & \textbf{223.2}\\
\hline
\end{tabular}}
\caption{Comparing aggregation approaches.}
\label{tab_supplemntary:encoding-approaches_YC2}
\end{center}
\end{wraptable}
Table~\ref{tab_supplemntary:encoding-approaches_YC2} shows that aggregating features has a poor performance. Moreover, using the middle output outperforms the other two aggregation techniques, as the model is enriching  the embedding of the anchor clip from its contextual neighbours.

\subsection{Results of Modality Context}
\label{sec:exp-modality-context}

\begin{figure}[t]
   \begin{minipage}{0.49\textwidth}
     \centering
     \includegraphics[width=\linewidth]{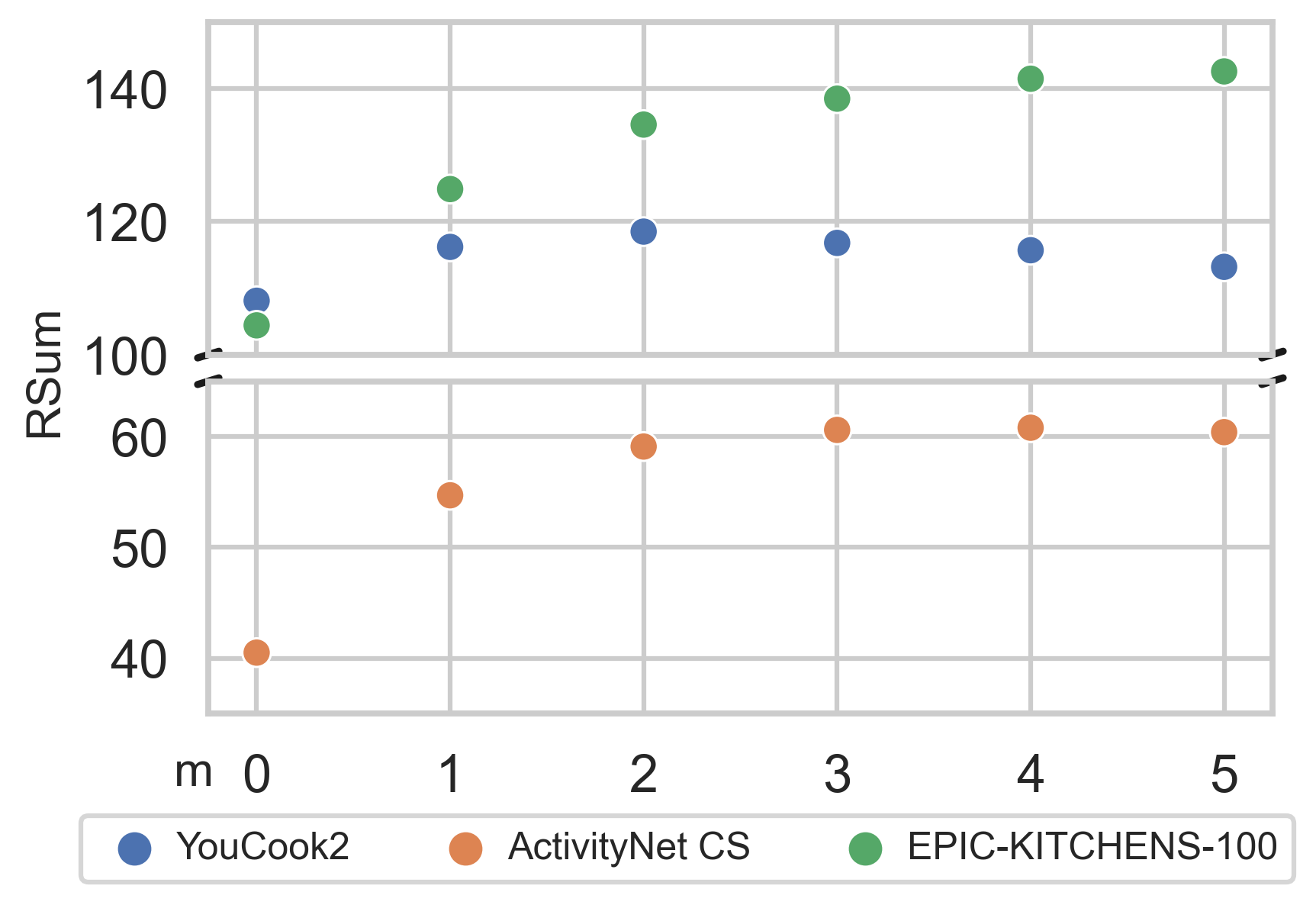}
     \caption{Analysis of temporal text context~(TC), reporting RSum in S2C.}
    \label{fig:fig_textcontext}
   \end{minipage}\hfill
   \begin{minipage}{0.49\textwidth}
     \centering
     \includegraphics[width=\linewidth]{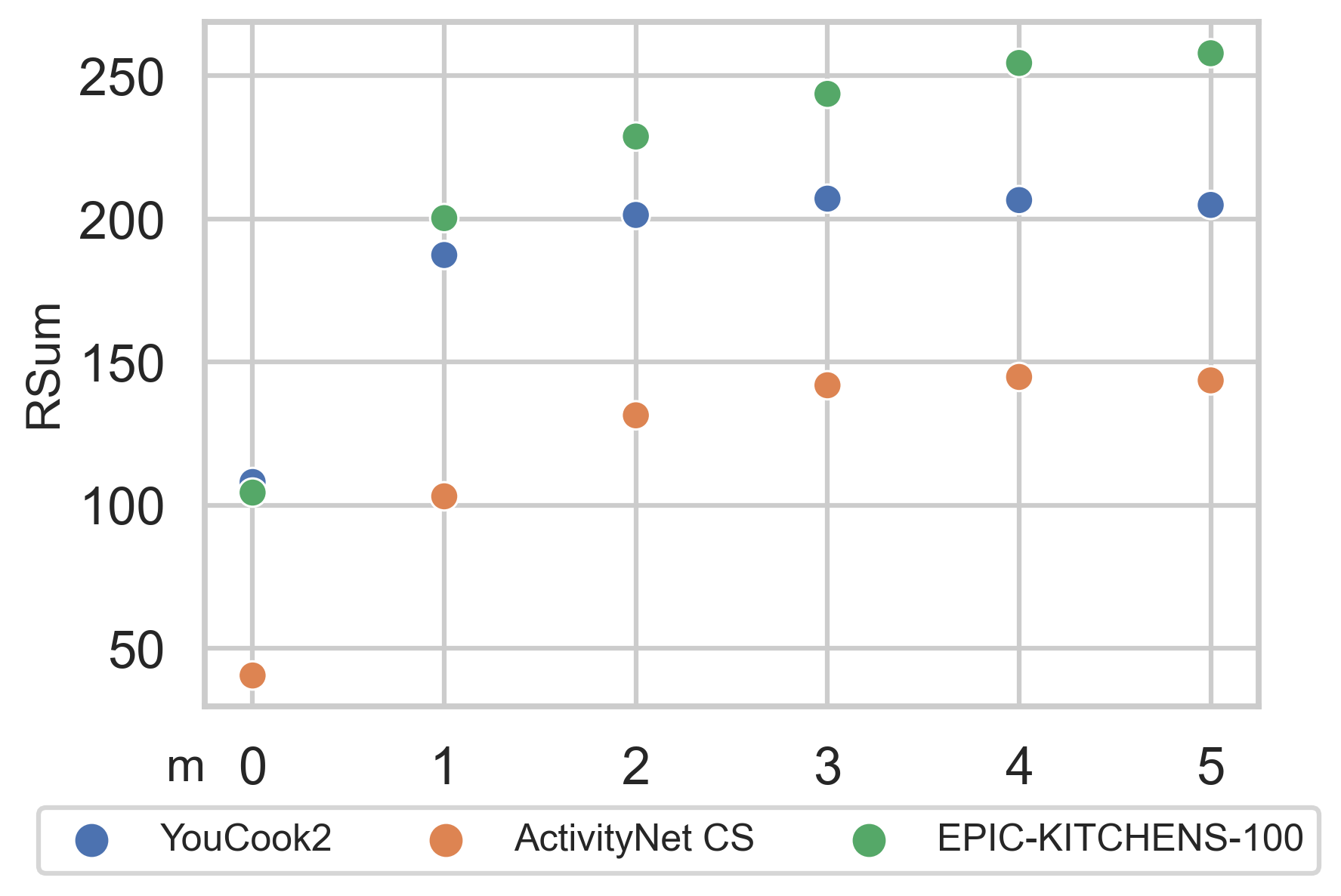}
     \caption{Analysis of temporal both context~(BC), reporting RSum in S2C.}
     \label{fig:fig_allcontext}
   \end{minipage}
\end{figure}

In Fig.~\ref{fig:fig_textcontext} and Fig.~\ref{fig:fig_allcontext} we provide comparable analysis as $m$ increases from no context ($m=0$) up to $m=5$, for text context (Fig.~\ref{fig:fig_textcontext}) and context in both modalities (Fig.~\ref{fig:fig_allcontext}). Results consistently demonstrate context to be helpful in both cases, $m=1$ outperforms $m=0$ by a large margin in every case. For some datasets, e.g.~ActivityNet CS, performance saturates and drops slightly for $m>3$. For long videos, e.g.~EPIC-KITCHENS-100, performance continues to improve with larger context.

\begin{figure*}[t]
    \centering
    \includegraphics[width=\linewidth]{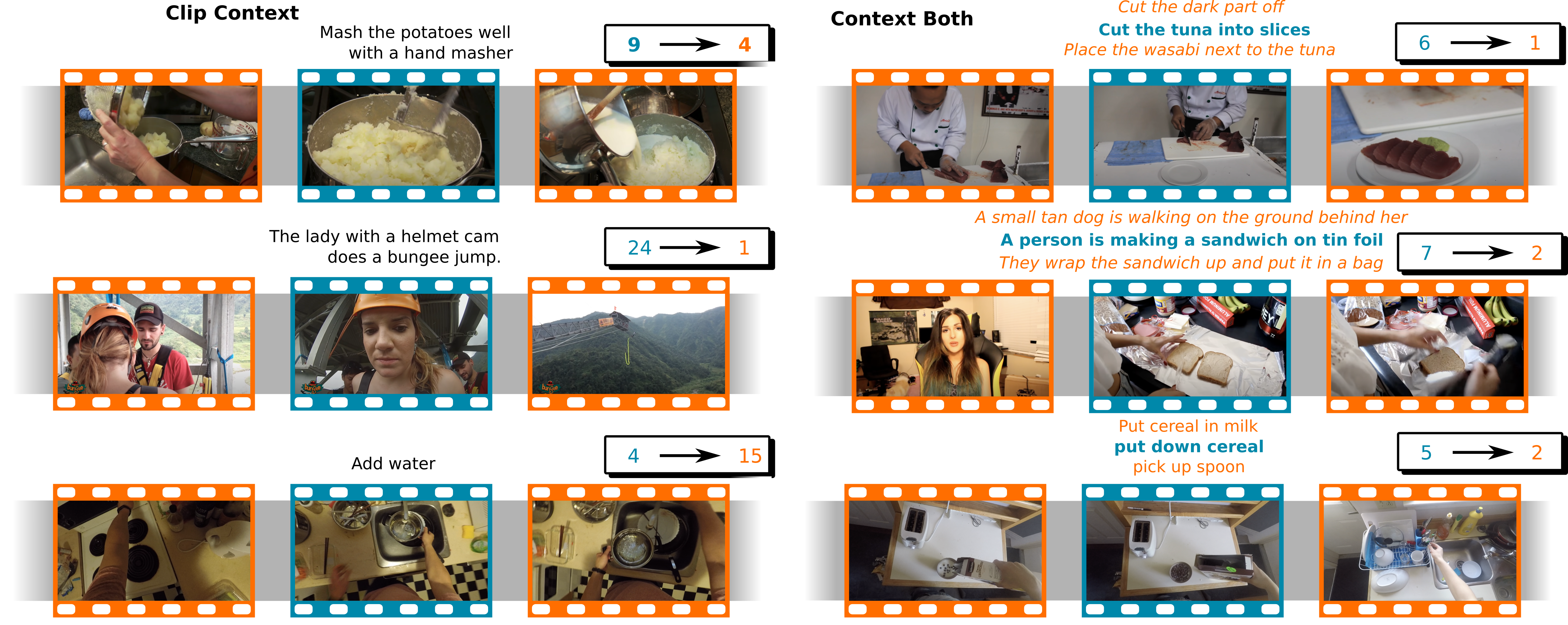}
    \vspace*{-12pt}
    \caption{Qualitative clip-to-sentence results for clip context (left) and context in both modalities (right) from 3 datasets: YouCook2 (top), ActivityNet CS (middle), and EPIC-KITCHENS-100 (bottom). The change in rank of retrieved video from no context~(\textcolor{cyan}{cyan}) to using context (\textcolor{orange}{orange}), e.g. from rank 9 to rank 4 when using context.}
    \label{fig:qual_results}
\vspace*{-12pt}
\end{figure*}

In Fig.~\ref{fig:qual_results} we show qualitative examples from models trained with clip context and context in both modalities. 
Clip context (left) shows additional visual context helps. For example, in row 1, the previous clip includes potatoes before being mashed---a more recognisable shape compared to their mashed state.
When using context in both modalities (right), these benefits are combined, leading to the model discriminating between difficult examples in which very similar clips are described.
\paragraph{Limitations.}\quad
We find that local context is certainly beneficial for Clip-Sentence retrieval, but also acknowledge there are cases in which it is detrimental.
An example of this is in Fig.~\ref{fig:qual_results} (row 3, left---``add water''), context drops the rank of the correct clip from 4 to 15. Studying the correct video, we note that neighbouring clips are not always related to the main action, i.e. ---``turn on the cooker''. The enriched clip representation is thus less similar to the query sentence.
In Fig~\ref{fig:rank_changes}, we also show specific words harmed by clip context. 


\section{Conclusions}
\vspace*{-4pt}

In this work, we introduce the notion of local temporal context, for clip-sentence cross-modal retrieval in long videos.
We propose an attention-based deep encoder, which we term Context Transformer (ConTra), that is trained using contrastive losses from the embedding space.
We demonstrate the impact of ConTra on individual modalities as well as both modalities in cross-modal retrieval.
We ablate our method to further show the benefit of each component.
Our results indicate, both qualitatively and by comparing to other approaches, that local context in retrieval decreases the ambiguity in clip-sentence retrieval on three video datasets. 
\vspace*{4pt}

\noindent \textbf{Acknowledgements} This work used public dataset and was supported by EPSRC UMPIRE (EP/T004991/1) and Visual AI (EP/T028572/1).
\clearpage

\bibliographystyle{splncs04}
\bibliography{egbib}
\clearpage
\title{ConTra: (Con)text (Tra)nsformer\\for Cross-Modal Video Retrieval \\ \vspace*{12pt} Supplementary Material}
\titlerunning{ConTra for Cross-Modal Video Retrieval - Supplementary}
%
\author{Adriano Fragomeni \qquad \quad Michael Wray \qquad \quad Dima Damen}
\authorrunning{A. Fragomeni \etal}
%
\institute{Department of Computer Science, University of Bristol, UK}
\maketitle              
%


In the supplementary material we first detail the tables corresponding to the figures in the main paper in Sec.~\ref{sec_tables_from_figures}. The supplementary also includes additional ablation experiments in Sec.~\ref{sec_supp_extra_ablation}.
 We then provide an analysis of the complexity of the proposed method in Sec.~\ref{sec_supp_complexity_analysis}.

\section{Tables corresponding to figures in the main paper}
\label{sec_tables_from_figures}

\noindent
Table~\ref{tab:context_video} details the results presented in Fig.~\textcolor{red}{3} and reports Recall at $K=\{1,5,10\}$ (R@K) and median rank (MR) for all the datasets when different lengths of local clip context are used.
\begin{table*}[h!]
\vspace{-1mm}
\resizebox{1\linewidth}{!}{%
\begin{tabular}{c|cccr|cccr|c||cccr|cccr|c||cccr|cccr|c}
& \multicolumn{9}{c||}{\textbf{YouCook2}}&\multicolumn{9}{|c||}{\textbf{ActivityNet CS}}&\multicolumn{9}{c}{\textbf{EPIC-KITCHENS-100}}\\ 
\cline{2-28}
& \multicolumn{4}{c|}{\textbf{Sentence-To-Clip}}& \multicolumn{4}{c|}{\textbf{Clip-To-Sentence}}& &\multicolumn{4}{c|}{\textbf{Sentence-To-Clip}}& \multicolumn{4}{c|}{\textbf{Clip-To-Sentence}}&&\multicolumn{4}{c|}{\textbf{Sentence-To-Clip}}& \multicolumn{4}{c|}{\textbf{Clip-To-Sentence}}&\\ 
\cline{1-28}
\multicolumn{1}{c|}{$m$} &\multicolumn{1}{c}{\smaller{R@1}} &\multicolumn{1}{c}{\smaller{R@5}}&\multicolumn{1}{c}{\smaller{R@10}} &\multicolumn{1}{r|}{\smaller{MR}} 
&\multicolumn{1}{c}{\smaller{R@1}} &\multicolumn{1}{c}{\smaller{R@5}}&\multicolumn{1}{c}{\smaller{R@10}} &\multicolumn{1}{r|}{\smaller{MR}}&\multicolumn{1}{c||}{\smaller{RSum}}&\multicolumn{1}{c}{\smaller{R@1}} &\multicolumn{1}{c}{\smaller{R@5}}&\multicolumn{1}{c}{\smaller{R@10}} &\multicolumn{1}{r|}{\smaller{MR}} 
&\multicolumn{1}{c}{\smaller{R@1}} &\multicolumn{1}{c}{\smaller{R@5}}&\multicolumn{1}{c}{\smaller{R@10}} &\multicolumn{1}{r|}{\smaller{MR}}&\multicolumn{1}{c||}{\smaller{RSum}}&\multicolumn{1}{c}{\smaller{R@1}} &\multicolumn{1}{c}{\smaller{R@5}}&\multicolumn{1}{c}{\smaller{R@10}} &\multicolumn{1}{r|}{\smaller{MR}} &\multicolumn{1}{c}{\smaller{R@1}} &\multicolumn{1}{c}{\smaller{R@5}}&\multicolumn{1}{c}{\smaller{R@10}} &\multicolumn{1}{r|}{\smaller{MR}}&\multicolumn{1}{c}{\smaller{RSum}} \\\hline
0 & 15.6 & 39.1 & 52.0 & 10 & 13.3 & 37.1 & 50.6 & 10 & 207.7 & 4.3 & 14.3 & 21.9 & 51 & 4.4 & 14.6 & 22.9 & 50 & 82.4 & 17.9 & 38.6 & 47.9 & 12 & 24.0 & 46.4 & 55.7 & 7 & 230.5\\
1 & 16.5 & 41.1 & 54.2 & \textbf{8} & 14.1 & 38.6 & 51.7 & 10 & 216.2 & 5.7 & 17.5 & 26.0 & 41 & 5.7 & 17.9 & 27.2 & 39 & 100.0 & 21.4 & 42.5 & 51.8 & 10 & 27.8 & 50.5 & 59.7 & \textbf{5} & 253.7\\
2 & \textbf{17.1} & 42.0 & 54.4 & \textbf{8} & \textbf{14.9} & 39.8 & 52.8 & \textbf{9} & 221.0 & 5.8 & 17.9 & 26.8 & 38 & 6.2 & 19.2 & 28.3 & 37 & 104.2 & 21.6 & 42.6 & 53.0 & 9 & 28.0 & 51.0 & 60.7 & \textbf{5} & 256.9\\
3 & 16.7 & \textbf{42.1} & \textbf{55.2} & \textbf{8} & 14.8 & 40.5 & 53.9 & \textbf{9} & \textbf{223.2} & \textbf{5.9} & 18.4 & \textbf{27.6} & \textbf{38} & \textbf{6.4} & 19.3 & \textbf{28.5} & 37 & \textbf{106.1} & 21.6 & 43.2 & \textbf{53.6} & \textbf{8} & \textbf{28.6} & 51.6 & \textbf{61.3} & \textbf{5} & 259.9\\
4 & 16.7 & 41.8 & 54.9 & \textbf{8} & 14.7 & 40.3 & \textbf{54.5} & \textbf{9} & 222.9 & \textbf{5.9} & \textbf{18.5} & 27.2 & \textbf{38} & \textbf{6.4} & \textbf{19.4} & 28.4 & \textbf{36} & 105.8 & \textbf{22.2} & 43.4 & 53.4 & 9 & 28.2 & \textbf{52.0} & 61.1 & \textbf{5} & \textbf{260.3}\\
5 &16.5 & \textbf{42.1} & 54.4 & \textbf{8} & 14.8 & \textbf{40.7} & 53.7 & \textbf{9} & 222.2 & 5.8 & 18.1 & 27.2 & \textbf{38} & 6.3 & 19.1 & 28.4 & \textbf{36} & 104.9 & 21.2 & \textbf{43.7} & 53.5 & \textbf{8} & 28.2 & 51.6 & 61.2 & \textbf{5} & 259.4\\\hline
\end{tabular}}
\caption{Analysis of the importance of temporal clip context (CC), reporting Recall $\uparrow$ and Median Rank $\downarrow$.}
\label{tab:context_video}
\vspace{-20pt}
\end{table*}

Table~\ref{tab:context_text} and Table~\ref{tab:both_context} are the extensions of Fig.~\textcolor{red}{7} and Fig.~\textcolor{red}{8} from the main paper, respectively, and report sentence-to-clip results for all the datasets when different lengths of local clip context are used.

\begin{table*}[h!]
\vspace{-12pt}
\centering
\begin{tabular}{c|cccr|c||cccr|c||cccr|c}

& \multicolumn{5}{c||}{\textbf{YouCook2 (S2C)}}&\multicolumn{5}{|c||}{\textbf{ActivityNet CS (S2C)}}&\multicolumn{5}{c}{\textbf{EPIC-KITCHENS-100 (S2C)}}\\ 
\cline{2-16}

\multicolumn{1}{c|}{$m$} &\multicolumn{1}{c}{\smaller{R@1}} &\multicolumn{1}{c}{\smaller{R@5}}&\multicolumn{1}{c}{\smaller{R@10}} &\multicolumn{1}{l|}{\smaller{MR}} 
&\multicolumn{1}{c||}{\smaller{RSum}}
&\multicolumn{1}{c}{\smaller{R@1}} &\multicolumn{1}{c}{\smaller{R@5}}&\multicolumn{1}{c}{\smaller{R@10}} &\multicolumn{1}{c|}{\smaller{MR}}&\multicolumn{1}{c||}{\smaller{RSum}} &\multicolumn{1}{c}{\smaller{R@1}} &\multicolumn{1}{c}{\smaller{R@5}}&\multicolumn{1}{c}{\smaller{R@10}} &\multicolumn{1}{l|}{\smaller{MR}}&\multicolumn{1}{c}{\smaller{RSum}} \\\hline

0 & 16.0 & 39.7 & 52.5 & 9 & 108.1& 4.3 & 14.3 & 21.9 & 51 & 40.5& 17.9 & 38.6 & 47.9 & 12 & 104.4\\
1 & 17.5 & 42.7 & 56.0 & 8 & 116.2& 6.0 & 19.5 & 29.2 & 30 & 54.7 &22.0 &46.6& 56.3 & 7 & 124.9   \\
2 & \textbf{17.7} & \textbf{43.9} & \textbf{56.9} & \textbf{7} & \textbf{118.5} & 6.7 & 21.1 & 31.3 & 27 & 59.1 & 24.1 & 50.1 & 60.4 & \textbf{5} &134.6 \\
3 & 17.3 & 42.7 & 56.8 & 8 & 116.8 & 6.9 & 21.7 & \textbf{32.0} & \textbf{26} & 60.6 & 25.6 & 51.2 & 61.7 & \textbf{5} & 138.5\\
4 & 17.1 & 42.8 & 55.8 & 8 & 115.7& \textbf{7.0} & \textbf{21.8} & \textbf{32.0} & \textbf{26} & \textbf{60.8} & \textbf{26.2} & 52.6 & 62.7 & \textbf{5} &  141.5\\
5 & 16.6 & 41.2 & 55.4 & 8 & 113.2 & \textbf{7.0} & 21.5 & 31.9 & \textbf{26} & 60.4 & 25.8 & \textbf{53.2} & \textbf{63.6} & \textbf{5} &  \textbf{142.6}\\\hline
\end{tabular}
\caption{Analysis of temporal context in text only for sentence-to-clip.}
\label{tab:context_text}
\end{table*}

\begin{table*}[h!]
\centering

\begin{tabular}{c|cccr|c||cccr|c||cccr|c}

& \multicolumn{5}{c||}{\textbf{YouCook2 (S2C)}}&\multicolumn{5}{c||}{\textbf{ActivityNet CS (S2C)}}&\multicolumn{5}{c}{\textbf{EPIC-KITCHENS-100 (S2C)}}\\ 
\cline{2-16}

\multicolumn{1}{c|}{$m$} &\multicolumn{1}{c}{\smaller{R@1}} &\multicolumn{1}{c}{\smaller{R@5}}&\multicolumn{1}{c}{\smaller{R@10}} &\multicolumn{1}{l|}{\smaller{MR}} 
&\multicolumn{1}{c||}{\smaller{RSum}}
&\multicolumn{1}{c}{\smaller{R@1}} &\multicolumn{1}{c}{\smaller{R@5}}&\multicolumn{1}{c}{\smaller{R@10}} &\multicolumn{1}{c|}{\smaller{MR}}&\multicolumn{1}{c||}{\smaller{RSum}} &\multicolumn{1}{c}{\smaller{R@1}} &\multicolumn{1}{c}{\smaller{R@5}}&\multicolumn{1}{c}{\smaller{R@10}} &\multicolumn{1}{l|}{\smaller{MR}}&\multicolumn{1}{c}{\smaller{RSum}} \\\hline

0 & 16.0 & 39.7 & 52.5 & 9 & 108.1 & 4.3 & 14.3 & 21.9 & 51 & 40.5& 17.9 & 38.6 & 47.9 & 12 & 104.4\\

1 & 40.1 & 68.5 & 78.8 & \textbf{2} & 187.4 & 16.4  & 37.3 & 49.4 & 11 & 103.1 & 48.0 & 72.5 & 79.8 & 2 & 200.3\\

2 & 45.5 & 73.7 & 82.2 & \textbf{2} & 201.4 & 22.7 & 48.1 & 60.6 & 6 & 131.4 & 60.6 & 81.3 & 86.9 & \textbf{1} &  228.8\\

3 & \textbf{47.0} & \textbf{75.4} & 84.7 & \textbf{2} &  \textbf{207.1} & 25.3 & 52.6 & 64.8 & \textbf{5} & 141.9& 67.0 & 85.9 & 90.8 & \textbf{1} & 243.7\\

4 & 46.4 & 75.3 & \textbf{84.9} & \textbf{2} & 206.6& \textbf{26.3} & \textbf{53.2} & \textbf{65.3} & \textbf{5} &\textbf{144.8} & 72.2 & 89.4 & 92.9 & \textbf{1} & 254.5 \\

5 & 46.3 & 74.7 & 83.9 & \textbf{2} & 204.9& 25.7 & \textbf{53.2} & 64.7 & \textbf{5} & 143.6 & \textbf{73.8} & \textbf{90.4} & \textbf{93.7} & \textbf{1} & \textbf{257.9}\\\hline
\end{tabular}
\caption{Analysis of temporal context in both text and video for sentence-to-clip.}
\label{tab:both_context}
\end{table*}

\section{Extra Ablation Studies}
\label{sec_supp_extra_ablation}

In this section, we present some additional experiments on all datasets. Additional ablation studies on YouCook2 are in Sec.~\ref{sec_supp_YC2}, ActivityNet Clip-Sentence in Sec.~\ref{sec_supp_ActNet}, EPIC-KITCHENS-100 in Sec.~\ref{sec_supp_Epic}, a further analysis on the neighbouring loss, $L_{NEI}$ is in Sec.~\ref{sec_supp_NEI_loss} followed by context vs clip length in Sec.~\ref{sec_context_analysis}.

\subsection{Ablations on YouCook2 (YC2)}
\label{sec_supp_YC2}

\noindent
\textbf{Number of Heads and Layers.}\quad
We study how the performance changes by varying the number of stacked encoder layers in clip and text transformers, and the number of heads per layer
in Table \ref{tab_supplemntary:number_layers_heads_YC2}. 
\begin{table}[t]
\vspace{-12pt}
\begin{center}

\begin{tabular}{cc|cc|cccc|cccc|c}

\multicolumn{2}{c|}{\textbf{\#Layers}}&\multicolumn{2}{c|}{\textbf{\#Heads}} &\multicolumn{4}{c|}{\textbf{Sentence-to-Clip}} & \multicolumn{4}{c|}{\textbf{Clip-to-Sentence}} & \\
\cline{1-12}

\multicolumn{1}{c}{V}&\multicolumn{1}{c|}{T}&\multicolumn{1}{c}{V}&\multicolumn{1}{c|}{T} &\multicolumn{1}{c}{R@1} &\multicolumn{1}{c}{R@5}&\multicolumn{1}{c}{R@10} &\multicolumn{1}{l|}{MR} 
&\multicolumn{1}{c}{R@1} &\multicolumn{1}{c}{R@5}&\multicolumn{1}{c}{R@10} &\multicolumn{1}{c|}{MR}&
\multicolumn{1}{c}{RSum} \\\hline

1 & 1 & 1 & 1 & \textbf{17.1} & 42.0 & \textbf{55.2} & \textbf{8} & 14.4 & 39.3 & 53.5 & \textbf{9} & 221.5\\
1 & 1 & 2 & 2 & 16.7 & \textbf{42.1} & \textbf{55.2} & \textbf{8} & 14.8 & \textbf{40.5} & \textbf{53.9} & \textbf{9} & \textbf{223.2}\\
1 & 1 & 4 & 4 & 16.5 & 41.7 & \textbf{55.2} & \textbf{8} & 14.6 & 39.9 & 53.8 & \textbf{9} & 221.7\\
1 & 1 & 8 & 8 & 16.7 & 41.3 & \textbf{55.2} & \textbf{8} & \textbf{15.2} & 39.9 & 53.6 & \textbf{9} & 221.9\\
1 & 1 & 16 & 16 & 16.4 & 41.6 & 55.0 & \textbf{8} & 14.8 & 40.0 & 53.5 & \textbf{9} & 221.3 \\\hline

2 & 1 & 2 & 2 & 16.1 & 41.1 & 54.9 & \textbf{8} & 14.5 & 39.3 & 52.9 & \textbf{9} & 218.8\\
1 & 2 & 2 & 2 & 16.1 & 41.1 & 53.8 & 9 & 14.7 & 39.8 & 52.9 & \textbf{9} & 218.4\\
2 & 2 & 2 & 2 & 16.2 & 40.3 & 54.2 & 9 & 14.5 & 39.4 & 52.3 & \textbf{9} & 216.9\\\hline

\end{tabular}

\caption{Different number of layers and heads for Video (V) and Text (T) transformer encoders (YC2).}
\vspace*{-35pt}
\label{tab_supplemntary:number_layers_heads_YC2}
\end{center}
\end{table}
We first fix the number of layers and change the number of heads in both encoders.
The best result is achieved when using only $2$ heads. Overall, we can assert that the method is robust to the number of heads, as performance varies only marginally when the number of heads is adjusted. 
Next, we fix the number of heads to $2$ -- our best result from above, and change the number of layers in our encoders, for both clip transformer and text transformer. The best performance is achieved using $1$ layer. 
Note that YouCook2 is the smallest dataset. We ablate the number of heads and layers also in Sec~\ref{sec_supp_Epic} for larger datasets.

\noindent
\textbf{Temperature Parameter.}\quad 
Several works \cite{DBLP:conf/cvpr/He0WXG20,DBLP:journals/corr/abs-2003-04298,DBLP:conf/cvpr/WuXYL18} proposed to set the temperature parameter $\tau = 0.07$.
\begin{table}[h!]
\vspace*{-12pt}
\begin{center}
\begin{tabular}{l|cccr|cccr|c}

&\multicolumn{4}{c|}{\textbf{Sentence-to-Clip}} & \multicolumn{4}{c|}{\textbf{Clip-to-Sentence}} & \\
\cline{2-9}

\multicolumn{1}{c|}{$\tau$} &\multicolumn{1}{c}{R@1} &\multicolumn{1}{c}{R@5}&\multicolumn{1}{c}{R@10} &\multicolumn{1}{l|}{MR} 
&\multicolumn{1}{c}{R@1} &\multicolumn{1}{c}{R@5}&\multicolumn{1}{c}{R@10} &\multicolumn{1}{c|}{MR}&
\multicolumn{1}{c}{RSum} \\\hline

0.05 & \textbf{17.2} & 41.9 & \textbf{55.6} & \textbf{8} & \textbf{15.1} & 40.4 & \textbf{54.7} & \textbf{8} & \textbf{224.9}\\
0.07 & 16.7 & \textbf{42.1} & 55.2 & \textbf{8} & 14.8 & \textbf{40.5} & 53.9 & 9 & 223.2\\
0.1 & 16.2 & 40.3 & 53.9 & 9 & 14.5 & 39.6 & 52.5 & 9 & 217.0\\
0.7 & 12.5 & 33.2 & 46.0 & 13 & 10.9 & 31.8 & 43.1 & 15 & 177.5\\
1.0 & 13.0 & 34.9 & 47.2 & 12 & 12.5 & 33.3 & 45.2 & 13 & 186.1\\\hline

\end{tabular}
\caption{Analysis of the performance varying $\tau$ in $L_{CML}$ and $L_{NEI}$ (YC2).}
\vspace*{-35pt}
\label{tab_supplementary:tau_YC2}

\end{center}
\end{table}
\noindent
We thus follow this in all our experiments.
Here, we test empirically this choice by varying $\tau$ as shown in Table \ref{tab_supplementary:tau_YC2}. 
Increasing the value of $\tau$, drops the performance of our model in both retrieval tasks. 
While higher, but comparable, results are achieved with $\tau = 0.05$, we keep the standard $\tau$ to remain directly comparable to other works that keep $\tau$ at $0.07$.

\noindent
\textbf{Loss weights.}\quad
In the main paper, our objective function is defined as follows:
\begin{equation}
L=\lambda_{CML}L_{CML} + \lambda_{NEI}L_{NEI} +\lambda_{UNI}L_{UNI}
\end{equation}
where $\lambda_{CML}=\lambda_{UNI}=\lambda_{NEI}=1$ showcasing that we outperform other methods without hyperparameter tuning. Table \ref{tab:weights_loss_clipcontext} shows the performance of ConTra on YouCook2 when different combination of weights are used.
By applying a Grid Search approach we are able to find the best combination of weights of the multiple loss empirically and improve the RSum of $7.6$ points when $\lambda_{CML}=1$, $\lambda_{UNI}=12$ and $\lambda_{NEI}=2$. 
The uniformity loss's weight has the largest value, i.e. $\lambda_{UNI}=12$, but this high value is similar to other works \cite{DBLP:conf/cvpr/ChunORKL21}. 
\begin{table}[h!]
\begin{center}
\vspace*{-12pt}
\begin{tabular}{ccc|cccr|cccr|c}
\multicolumn{3}{c|}{Weights Loss} &\multicolumn{4}{c|}{\textbf{Sentence-to-Clip}}&\multicolumn{4}{c|}{\textbf{Clip-to-Sentence}} &\\
\cline{1-11}

\multicolumn{1}{c}{$\lambda_{CML}$}& \multicolumn{1}{c}{$\lambda_{UNI}$}& \multicolumn{1}{c|}{$\lambda_{NEI}$}&\multicolumn{1}{c}{R@1} &\multicolumn{1}{c}{R@5}&\multicolumn{1}{c}{R@10} &\multicolumn{1}{r|}{MR}&\multicolumn{1}{c}{R@1} &\multicolumn{1}{c}{R@5}&\multicolumn{1}{c}{R@10} &\multicolumn{1}{r|}{MR}&
\multicolumn{1}{c}{RSum} \\\hline

1& 1 & 1 & 16.7 & 42.1 & 55.2 & 8 & 14.8 & 40.5 & 53.9 & 9 & 223.2\\\hline
2 & 1 & 1 & 16.6 & 40.8 & 55.1 & 8 & 14.5 & 39.5 & 53.5 & 9 & 220.0 \\\hline
1  & 1 & 2 & 16.9 & 42.7 & 55.6 & 8 & 14.8 & 39.9 & 54.2 & 9 & 224.1\\
1  & 1 & 5 & 16.1 & 40.4 & 54.3 & 9 & 14.6 & 38.7 & 52.4 & 9 & 216.5\\\hline
1  & 2 & 1 & \textbf{17.3} & 42.1 & 55.3 & 8 & 15.1 & 40.4 & 54.3 & 9 & 224.5\\
1  & 5 & 1 & 17.1 & 42.8 & 55.7 & 8 & 15.5 & 41.1 & 54.8 & \textbf{8} & 227.0\\
1  & 10 & 1 & 16.7 & 43.2 & 56.2 & 8 & 15.7 & 41.6 & 55.9 & \textbf{8} & 229.3\\
1  & 12 & 1 & 17.0 & \textbf{43.4} & 56.1 & 8 & 15.6 & \textbf{42.2} & 55.9 & \textbf{8} & 230.2\\
1  & 15 & 1 & 17.1 & 42.9 & 56.0 & 8 & 15.6 & 41.8 & 55.6 & \textbf{8} & 229.0\\\hline
1  & 12 & 2 & 16.9 & 43.3 & \textbf{56.8} & \textbf{7} & \textbf{15.9} & 41.7 & \textbf{56.2} & \textbf{8} & \textbf{230.8}\\\hline

\end{tabular}

\caption{Ablation of loss weights in clip context scenario (YC2).}
\label{tab:weights_loss_clipcontext}
\vspace*{-35pt}
\end{center}
\end{table}

\noindent
\textbf{Batch Size.}\quad We vary the batch size in Table~\ref{tab:losbatchsize}. 
The overall performance of the model increases with the batch size, showing the importance of having varied negatives per batch when using the NCE loss.
Given memory limits, we were unable to increase the batch size further, but we anticipate diminishing returns above 512.
\noindent
\begin{table}[h!]
\begin{center}

\begin{tabular}{c|cccc|cccc|c}
&\multicolumn{4}{c|}{\textbf{Sentence-to-Clip}} & \multicolumn{4}{c|}{\textbf{Clip-to-Sentence}} & \\
\cline{2-9}
\multicolumn{1}{c|}{$B$} &\multicolumn{1}{c}{R@1} &\multicolumn{1}{c}{R@5}&\multicolumn{1}{c}{R@10} &\multicolumn{1}{l|}{MR} 
&\multicolumn{1}{c}{R@1} &\multicolumn{1}{c}{R@5}&\multicolumn{1}{c}{R@10} &\multicolumn{1}{c|}{MR}&
\multicolumn{1}{c}{RSum} \\\hline
64  & 15.8 & 41.2 & 54.6 & \textbf{8} & 13.8 & 39.5 & 52.4 & \textbf{9} & 217.3\\
128  & 16.4 & 41.7 & 55.1 & \textbf{8} & 14.2 & 40.0 & 53.1 & \textbf{9}  & 220.5\\
256  & \textbf{16.8} & 41.9& 54.7 & \textbf{8} & 14.4 & 39.5 & 52.9 & \textbf{9}  & 220.2\\
512 & 16.7 & \textbf{42.1} & \textbf{55.2} & \textbf{8} & \textbf{14.8} & \textbf{40.5} & \textbf{53.9} & \textbf{9} & \textbf{223.2}\\\hline
\end{tabular}
\caption{Analysis of the performance varying the batch size, $B$, (YC2).}

\label{tab:losbatchsize}
\end{center}
\end{table}

\begin{table*}[h!]
\begin{center}

\begin{tabular}{cc|cccr|c}
\multicolumn{1}{c}{Shared}& \multicolumn{1}{c|}{Positional} &\multicolumn{4}{c|}{\textbf{Sentence-to-Clip}}\\ 
\cline{3-7}

\multicolumn{1}{c}{Weights}& \multicolumn{1}{c|}{Encoding}  &\multicolumn{1}{c}{R@1} &\multicolumn{1}{c}{R@5}&\multicolumn{1}{c}{R@10} &\multicolumn{1}{l|}{MR}&
\multicolumn{1}{c}{RSum} \\\hline

\checkmark & shared & 47.2 & 74.8  & 83.7 & \textbf{2} & 205.7\\
\checkmark & distinct & 47.1 & 74.9 & 84.2 & \textbf{2} & 206.2\\
$ \times $ & shared & \textbf{47.3} & 74.8 & 83.5 & \textbf{2} & 205.6\\
$ \times $ & $ \times $ & 46.1 & 74.5 & 83.9 & \textbf{2} & 204.5\\
$ \times $ & distinct & 47.0 & \textbf{75.4} & \textbf{84.7} & \textbf{2} &  \textbf{207.1}\\\hline

\end{tabular}
\caption{Ablation of modality weights and pos. encoding (YC2).}
\label{tab_supplementary:both_context_positionalencoding}
\end{center}
\end{table*}



\noindent
\textbf{Shared Weights and Positional Encoding.}\quad We present additional results on YouCook2  for sentence-to-clip when utilising context in both modalities. We ablate the choice of weights and position encodings in both clip and text transformers. Table~\ref{tab_supplementary:both_context_positionalencoding} compares these results with shared/distinct weights and positional encodings. We also evaluate removing the positional encoding. Using different weights and distinct encodings between modalities achieves the best performance.

\subsection{Ablations on ActivityNet CS}
\label{sec_supp_ActNet}
\noindent
\textbf{Loss Function.}\quad 
In Table \ref{tab_supplementary:loss_ActNet} we show the improvement given by all the terms of our objective function. 
We obtain similar results to those we report in Table 6 in the main paper, where the our neighbouring loss $L_{NEI}$ helps the model to better distinguish clips compared to the standard hard mining approach proposed in \cite{DBLP:conf/bmvc/FaghriFKF18}.

\begin{table}[h!]
\vspace{-12pt}
\begin{center}
\begin{tabular}{l|cccr|cccr|c}
&\multicolumn{4}{c|}{\textbf{Sentence-to-Clip}} & \multicolumn{4}{c|}{\textbf{Clip-to-Sentence}} & \\
\cline{2-9}
\multicolumn{1}{l|}{Loss}
&\multicolumn{1}{c}{\smaller{R@1}} &\multicolumn{1}{c}{\smaller{R@5}}&\multicolumn{1}{c}{\smaller{R@10}} &\multicolumn{1}{l|}{\smaller{MR}} 
&\multicolumn{1}{c}{\smaller{R@1}} &\multicolumn{1}{c}{\smaller{R@5}}&\multicolumn{1}{c}{\smaller{R@10}} &\multicolumn{1}{c|}{\smaller{MR}}&
\multicolumn{1}{c}{\smaller{RSum}} \\\hline
$L_{NEI}$& 2.6 & 9.3 & 14.8& 96& 2.5& 9.3& 15.0 & 96& 53.5 \\
$L_{CML}$& 5.0& 16.3& 24.9& 43 & 5.6& 17.3& 26.2& 42& 95.3\\ 
$L_{CML}$+$L_{HardMining}$ & 5.1 & 16.5& 25.0& 43& 5.7& 17.5& 26.4& 41& 96.2\\
$L_{CML}$+$L_{NEI}$& \textbf{5.9} & 18.1 & 27.2 & \textbf{38} & 6.1 & 19.0 & 28.1 & \textbf{37} & 104.4\\
$L_{CML}$+$L_{NEI}$+$L_{UNI}$& \textbf{5.9} & \textbf{18.4} & \textbf{27.6} & \textbf{38} & \textbf{6.4} & \textbf{19.3} & \textbf{28.5} & \textbf{37} & \textbf{106.1} \\ \hline
\end{tabular}
\caption{Ablation of loss function (ActivityNet CS).}
\vspace*{-35pt}
\label{tab_supplementary:loss_ActNet}
\end{center}
\end{table}

\subsection{Experiments on EPIC-KITCHENS-100}
\label{sec_supp_Epic}

\noindent
\textbf{Loss Function.}\quad Similarly, Table \ref{tab_supplementary:loss_Epic} illustrates the performance of our model on EPIC-KITCHENS-100 when changing the objective function. 
As with the other datasets, the neighbouring loss $L_{NEI}$ works better than $L_{HardMining}$ and justifies all the terms of our objective function, i.e. $L_{NEI}$ and $L_{UNI}$.

\begin{table}[h!]
\begin{center}
\begin{tabular}{l|cccr|cccr|c}
&\multicolumn{4}{c|}{\textbf{Sentence-to-Clip}} & \multicolumn{4}{c|}{\textbf{Clip-to-Sentence}} & \\
\cline{2-9}
\multicolumn{1}{l|}{Loss}
&\multicolumn{1}{c}{\smaller{R@1}} &\multicolumn{1}{c}{\smaller{R@5}}&\multicolumn{1}{c}{\smaller{R@10}} &\multicolumn{1}{l|}{\smaller{MR}} 
&\multicolumn{1}{c}{\smaller{R@1}} &\multicolumn{1}{c}{\smaller{R@5}}&\multicolumn{1}{c}{\smaller{R@10}} &\multicolumn{1}{c|}{\smaller{MR}}&
\multicolumn{1}{c}{\smaller{RSum}} \\\hline
$L_{NEI}$
& 5.6 & 15.8 & 23.3 & 61 & 9.4 & 28.1 & 37.8 & 23 & 120.0\\
$L_{CML}$
& 21.7 & 42.4 & 51.6 & \textbf{9} & 28.1 & 50.8 & 60.2 & \textbf{5} & 254.8\\ 
$L_{CML}$+$L_{HardMining}$ & 20.9 & 42.4& 51.6& \textbf{9}& 27.8& 51.4& 60.8& \textbf{5}& 254.9\\
$L_{CML}$+$L_{NEI}$& 21.1 & 43.0 & 52.9 & \textbf{9} & 27.3 & 51.3 & 60.9 & \textbf{5} & 256.5\\
$L_{CML}$+$L_{NEI}$+$L_{UNI}$& \textbf{22.2} & \textbf{43.4} & \textbf{53.4} & \textbf{9} & \textbf{28.2} & \textbf{52.0} & \textbf{61.1} & \textbf{5} & \textbf{260.3} \\ \hline
\end{tabular}
\caption{Ablation of loss function (EPIC-KITCHENS-100).}
\vspace*{-20pt}
\label{tab_supplementary:loss_Epic}
\end{center}
\end{table}

\begin{table}[h!]
\begin{center}
\begin{tabular}{l|cccr|cccr|c}

&\multicolumn{4}{c|}{\textbf{Sentence-to-Clip}} & \multicolumn{4}{c|}{\textbf{Clip-to-Sentence}} & \\
\cline{2-9}

\multicolumn{1}{c|}{$B$} &\multicolumn{1}{c}{R@1} &\multicolumn{1}{c}{R@5}&\multicolumn{1}{c}{R@10} &\multicolumn{1}{l|}{MR} 
&\multicolumn{1}{c}{R@1} &\multicolumn{1}{c}{R@5}&\multicolumn{1}{c}{R@10} &\multicolumn{1}{c|}{MR}&
\multicolumn{1}{c}{RSum} \\\hline

64  & 19.0 & 41.0 & 50.6 & 10 & 25.7 & 48.9 & 58.6 & 6 & 243.8\\
128  & 20.7& 42.4 & 52.1 & \textbf{9} & 27.3 & 51.0 & 60.7 & \textbf{5} & 254.2\\
256  & 21.6& 43.1 & 52.8 & \textbf{9} & \textbf{28.4} & \textbf{52.0} & \textbf{61.6} & \textbf{5} & 259.5\\
512  & \textbf{22.2} & \textbf{43.4} & \textbf{53.4} & \textbf{9} & 28.2 & \textbf{52.0} & 61.1 & \textbf{5} & \textbf{260.3} \\\hline

\end{tabular}
\caption{Analysis of the performance varying the batch size, $B$ (EPIC-KITCHENS-100).}
\vspace*{-20pt}
\label{tab_supplementary:batchsize_Epic}

\end{center}
\end{table}
\noindent
\textbf{Batch Size.}\quad
Similar to the results obtained in Table~\ref{tab:losbatchsize} on YouCook2, increasing the size of the batch boosts the performance of our model also on EPIC-KITCHENS-100 as highlighted in Table \ref{tab_supplementary:batchsize_Epic}.

\noindent
\textbf{Number of Heads and Layers.}\quad
\begin{table}[h!]
\begin{center}

\begin{tabular}{cc|cc|cccr|cccr|c}

\multicolumn{2}{c|}{\textbf{\#Layers}}&\multicolumn{2}{c|}{\textbf{\#Heads}} &\multicolumn{4}{c|}{\textbf{Sentence-to-Clip}} & \multicolumn{4}{c|}{\textbf{Clip-to-Sentence}} & \\
\cline{1-12}

\multicolumn{1}{c}{V}&\multicolumn{1}{c|}{T}&\multicolumn{1}{c}{V}&\multicolumn{1}{c|}{T} &\multicolumn{1}{c}{R@1} &\multicolumn{1}{c}{R@5}&\multicolumn{1}{c}{R@10} &\multicolumn{1}{r|}{MR} 
&\multicolumn{1}{c}{R@1} &\multicolumn{1}{c}{R@5}&\multicolumn{1}{c}{R@10} &\multicolumn{1}{r|}{MR}&
\multicolumn{1}{c}{RSum} \\\hline

2 & 2 & 1 & 1 & 20.7 & 42.1 & 52.4 & \textbf{9} & 27.3 & 51.1 & 60.4 &\textbf{5}  & 254.0\\
2 & 2 & 2 & 2& 21.3 & 43.0 & 52.9 & \textbf{9} & 27.7 & 51.1 & 60.4 & \textbf{5} & 256.4\\
2 & 2 & 4 & 4 & 21.5 & 43.0 & 53.1 & \textbf{9} & 28.3 & 51.8 & 60.8 & \textbf{5}  & 258.5\\
2 & 2 & 8 & 8 & \textbf{22.2} & \textbf{43.4} & \textbf{53.4} & \textbf{9} & 28.2 & \textbf{52.0} & 61.1 & \textbf{5} &\textbf{260.3} \\
2 & 2 & 16 & 16 & 21.4 & \textbf{43.4} & 53.1 & \textbf{9} & 28.2 & \textbf{52.0} & \textbf{61.6} & \textbf{5} & 259.7\\\hline

1 & 1 & 8 & 8 & 20.4 & 41.0 & 50.9 & 10 & 26.1 & 50.2 & 60.2 & \textbf{5} & 249.1\\
2 & 1 & 8 & 8 & 21.4 & 41.8 & 52.3 & \textbf{9} & \textbf{28.6} & 51.5 & 61.3 & \textbf{5} & 256.9\\
1 & 2 & 8 & 8 & 20.8 & 41.0 & 51.6 & 10 & 26.0 & 50.0 & 60.2 & 6 & 249.6\\\hline

\end{tabular}
\caption{Different number of heads and layers (EPIC-KITCHENS-100) for Video (V) and Text (T) transformer encoders.}
\vspace*{-35pt}
\label{tab_supplemntary:number_layers_heads_Epic}
\end{center}
\end{table}
\noindent
Table \ref{tab_supplemntary:number_layers_heads_Epic} shows how the performance changes by varying the number of stacked encoder layers and the number of heads per layer. We first fix the number of layers and change the number of heads in both encoders.
We achieve the best result when using $8$ heads. Overall, we can see that the method performs better when increasing the number of heads. This can be explained by the larger size of EPIC-KITCHENS-100 compared to YouCook2. Then, we fix the number of heads to $8$ and change the number of stacked encoder layers. The best performance remains using $2$ layer. We use the same number of layers and heads for ActivityNet Clip-Sentence.

\subsection{Neighbouring Loss per Dataset}
\label{sec_supp_NEI_loss}
\begin{figure*}
\vspace{-12pt}
\centering
\includegraphics[width=\linewidth]{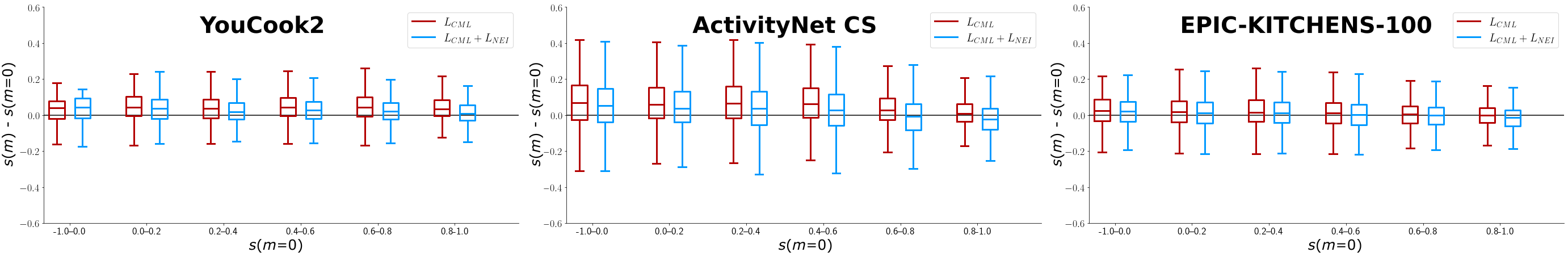}
\caption{Expanded version of Fig.~\textcolor{red}{6} from the main paper showing the comparison between similarities to neighbouring clips, $s(m)-s(m=0)$, with and without using $L_{NEI}$.}
\label{fig:supp_neigh_loss_all}
\end{figure*}

We provide an expanded version of Fig.~\textcolor{red}{6} from the main paper in Fig.~\ref{fig:supp_neigh_loss_all} broken down per dataset. Overall, the individual plots follow the results of Fig. 6 in the main paper. In each dataset we can see that $L_{NEI}$ has an important role, the similarity between neighbouring clips $j+1$ and the sentence $j$ tends to decrease for clips and sentences close in space (i.e. higher cosine similarity on the x-axis). 

\subsection{Context Length w.r.t. Clip Length}
\label{sec_context_analysis}

\begin{figure*}[h!]
\vspace{-12pt}
\centering
\includegraphics[width=\linewidth]{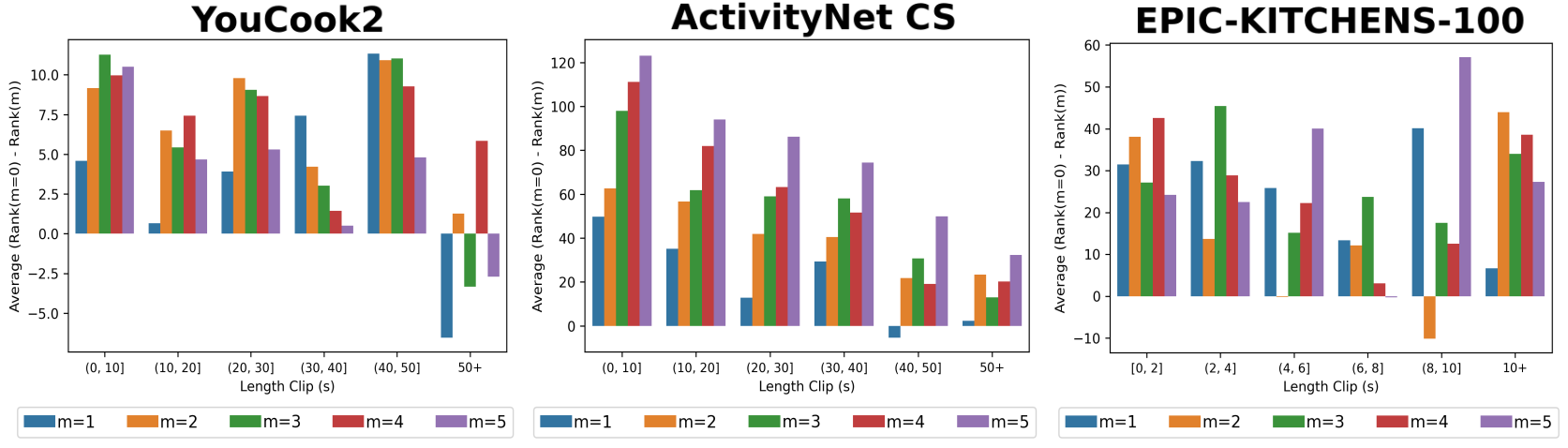}
\caption{Analysis of the improvements in the rank position w.r.t the length of the clip}
\label{fig:cliplength_analysis}
\vspace{-20pt}
\end{figure*}

We study the relation between context length and clip length by analysing the average improvement of the rank position of all the clips of a certain length, as shown in Fig.~\ref{fig:cliplength_analysis}.
We bin all the clips in the test set based on their length considering small intervals for dense datasets. We then calculate the average difference
between the rank position when $m=0$ and $m>0$. In all the datasets, short clips benefits the most from local clip context when $m>1$. 
Long clips demonstrate a different behaviour. In EPIC-KITCHENS-100 and ActivityNet CS, long context helps the most.
An interesting behaviour can be highlighted for ActivityNet CS. Although this dataset has the smallest number of clips per video on average, we see that the largest improvement of rank position when $m=5$. A possible explanation can be that in ActivityNet CS, some clips are as long as the entire video so using context adds fine-grained information that can help the model to retrieve these clips. 

\section{Complexity Analysis}
\label{sec_supp_complexity_analysis}
We present a computational complexity analysis of ConTra in Table~\ref{tab_supplementary:Complexity_clipcontex}, Table~\ref{tab_supplementary:Complexity_textcontext} and Table~\ref{tab_supplementary:Complexity_bothcontext} for local clip context, text context and when clip and text contexts are used simultaneously, respectively. 
\begin{table*}[h!]
\centering
\begin{tabular}{c|ccc|ccc|ccc}

& \multicolumn{3}{c|}{\textbf{YouCook2}}&\multicolumn{3}{c|}{\textbf{ActivityNet CS}}&\multicolumn{3}{c}{\textbf{EPIC-KITCHENS-100}}\\ 
\cline{2-10}

\multicolumn{1}{c|}{$m$} &\multicolumn{1}{c}{\smaller{\#Param.(M)}} &\multicolumn{1}{c}{\smaller{Flops(G)}}&\multicolumn{1}{c|}{\smaller{RSum}}
&\multicolumn{1}{c}{\smaller{\#Param.(M)}} &\multicolumn{1}{c}{\smaller{Flops(G)}}&\multicolumn{1}{c|}{\smaller{RSum}}
&\multicolumn{1}{c}{\smaller{\#Param.(M)}} &\multicolumn{1}{c}{\smaller{Flops(G)}}&\multicolumn{1}{c}{\smaller{RSum}}\\\hline

0& 9.549 & 0.04 & 207.7 & 8.675 & 0.02 & 82.4 & 16.903 & 0.06 & 230.5\\

1& 9.550 & 0.08 & 216.2 & 8.676 & 0.06 & 100.0 & 16.904 & 0.12 &253.7\\

2& 9.551 & 0.10 & 221.0 & 8.677 & 0.10 & 104.2 & 16.905 & 0.20 &256.9\\

3& 9.552 & 0.14 & 223.2 & 8.677 & 0.14 & 106.1 & 16.906 & 0.26 &259.9\\

4& 9.553 & 0.16 & 222.9 & 8.678 & 0.16 & 105.8 & 16.907 & 0.32 &260.3\\

5& 9.554 & 0.20 & 222.2 & 8.679 & 0.20 & 104.9 & 16.908 & 0.38 &259.4\\\hline
\end{tabular}

\caption{Analysis of Complexity (clip context).}
\label{tab_supplementary:Complexity_clipcontex}
\end{table*}

\begin{table*}[h!]
\centering
\begin{tabular}{c|ccc|ccc|ccc}

& \multicolumn{3}{c|}{\textbf{YouCook2 (S2C)}}&\multicolumn{3}{c|}{\textbf{ActivityNet CS (S2C)}}&\multicolumn{3}{c}{\textbf{EPIC-KITCHENS-100 (S2C)}}\\ 
\cline{2-10}

\multicolumn{1}{c|}{$m$} &\multicolumn{1}{c}{\smaller{\#Param.(M)}} &\multicolumn{1}{c}{\smaller{Flops(G)}}&\multicolumn{1}{c|}{\smaller{RSum}}
&\multicolumn{1}{c}{\smaller{\#Param.(M)}} &\multicolumn{1}{c}{\smaller{Flops(G)}}&\multicolumn{1}{c|}{\smaller{RSum}}
&\multicolumn{1}{c}{\smaller{\#Param.(M)}} &\multicolumn{1}{c}{\smaller{Flops(G)}}&\multicolumn{1}{c}{\smaller{RSum}}\\\hline

0& 9.549 & 0.04 & 108.1 & 8.675 & 0.02 & 40.5 & 16.903 & 0.06 & 104.4\\

1& 9.550 & 0.18 & 116.2 & 8.676 & 0.06 & 54.7 & 16,904 & 0.22 & 124.9\\

2& 9.551 & 0.28 & 118.5 & 8.677 & 0.10 & 59.1 & 16,905 & 0.36 &134.6\\

3& 9.552 & 0.40 & 116.8 & 8.677 & 0.14 & 60.6 & 16,906 & 0.50 &138.5\\

4& 9,553 & 0.52 & 115.7 & 8.678 & 0.16 & 60.8 & 16,907 & 0.64 &141.5\\

5& 9,554 & 0.62 & 113.2 & 8.679 & 0.20 & 60.4 & 16,908 & 0.78 &142.6\\\hline
\end{tabular}
\caption{Analysis of Complexity text context.}
\label{tab_supplementary:Complexity_textcontext}
\end{table*}

\begin{table*}[h!]
\centering

\begin{tabular}{c|ccc|ccc|ccc}

& \multicolumn{3}{c|}{\textbf{YouCook2 (S2C)}}&\multicolumn{3}{c|}{\textbf{ActivityNet CS (S2C)}}&\multicolumn{3}{c}{\textbf{EPIC-KITCHENS-100 (S2C)}}\\ 
\cline{2-10}

\multicolumn{1}{c|}{$m$} &\multicolumn{1}{c}{\smaller{\#Param.(M)}} &\multicolumn{1}{c}{\smaller{Flops(G)}}&\multicolumn{1}{c|}{\smaller{RSum}}
&\multicolumn{1}{c}{\smaller{\#Param.(M)}} &\multicolumn{1}{c}{\smaller{Flops(G)}}&\multicolumn{1}{c|}{\smaller{RSum}}
&\multicolumn{1}{c}{\smaller{\#Param.(M)}} &\multicolumn{1}{c}{\smaller{Flops(G)}}&\multicolumn{1}{c}{\smaller{RSum}}\\\hline

0& 9.549 & 0.04 & 108.1 & 8.675 & 0.02 & 40.5 & 16.903 & 0.06 & 104.4 \\

1& 9.551 & 0.22 & 187.4 & 8.677 & 0.10 & 103.1 & 16.905 & 0.30 & 200.3\\

2& 9.553 & 0.36 & 201.4 & 8.678 & 0.18 &131.4 & 16.907 & 0.50 & 228.8\\

3& 9.555 & 0.50 & 207.1 & 8.680 & 0.24 & 141.9 & 16.909 & 0.70 &243.7\\

4& 9.558 & 0.64 & 206.6 & 8.681 & 0.32 & 144.8 & 16.911 & 0.90 &254.5\\

5& 9.560 & 0.78 & 204.9 & 8.683 & 0.38 & 143.6 & 16.913 &1.10 &257.9\\\hline
\end{tabular}
\caption{Analysis of Complexity both context.}
\label{tab_supplementary:Complexity_bothcontext}
\end{table*}

For each dataset and each value of $m$, we report the number of trainable parameters, the FLOPS in training, and the performance of the model, i.e. total RSum in Table~\ref{tab_supplementary:Complexity_clipcontex} and RSum$_{S2C}$ in Table~\ref{tab_supplementary:Complexity_textcontext} and Table~\ref{tab_supplementary:Complexity_bothcontext}.
As noted in the tables, increased context only increases the number of parameters slightly, but require more GFlops.
As expected, using context in both modalities is the most computationally expensive setting as shown in Table~\ref{tab_supplementary:Complexity_bothcontext}.

In general, the performance tends to be the comparable or drops marginally for $m>3$ over all datasets/scenarios. We can argue that using a $1 \le m\leq3$ is a good trade-off between performance and computational complexity.
\clearpage
\end{document}